\documentclass{article}

\usepackage{microtype}
\usepackage{graphicx}
\usepackage{subfigure}
\usepackage{booktabs} 

\usepackage{hyperref}

\usepackage[accepted]{icml2024}

\usepackage{amsmath}
\usepackage{amssymb}
\usepackage{mathtools}
\usepackage{amsthm}

\usepackage[capitalize,noabbrev]{cleveref}

\newtheorem{definition}{Definition}
\newtheorem{lemma}{Lemma}
\newtheorem{proposition}{Proposition}
\usepackage[textsize=tiny]{todonotes}

\PassOptionsToPackage{numbers, compress}{natbib}
\usepackage{amsmath}
\usepackage{microtype}
\usepackage{graphicx}
\usepackage{subfigure}
\usepackage{hyperref}
\hypersetup{
    colorlinks=true,
    linkcolor=blue,
    filecolor=magenta,      
    urlcolor=cyan,
    citecolor=black
}
\usepackage{booktabs} 
\usepackage{amsfonts}
\usepackage{bm}
\usepackage{threeparttable}
\usepackage{tablefootnote}

\usepackage{lipsum}
\usepackage{comment}

\usepackage{multirow}
\usepackage{amsmath}

\usepackage{hyperref}

\usepackage{enumitem}

\usepackage{mathtools}

\usepackage{wrapfig}

\usepackage{listings}

\usepackage[english]{babel}

\usepackage{listings}
\usepackage{color}

\definecolor{mygreen}{rgb}{0,0.6,0}
\definecolor{mygray}{rgb}{0.5,0.5,0.5}
\definecolor{mymauve}{rgb}{0.58,0,0.82}

\usepackage{marginnote}
\usepackage{soul} 

\newcommand{\ignore}[1]{}



\usepackage{amsmath,amsfonts,bm}

\def\eqref#1{equation~\ref{#1}}

\def\1{\bm{1}}

\DeclareMathAlphabet{\mathsfit}{\encodingdefault}{\sfdefault}{m}{sl}
\SetMathAlphabet{\mathsfit}{bold}{\encodingdefault}{\sfdefault}{bx}{n}
\def \bq {{\bm q}}
\def \bk {{\bm k}}

\def \bx {{\bm x}}
\def \bz {{\bm z}}

\def \bv {{\bm v}}

\def \bu {{\bm u}}
\def \bff {{\bm f}}

\def \bQ {{\mathbf Q}}
\def \bK {{\mathbf K}}
\def \bV {{\mathbf V}}
\def \bW {{\mathbf W}}

\def \bX {{\mathbf X}}

\def \bI {{\mathbf I}}

\usepackage{amsmath}

\DeclareMathOperator*{\argmin}{arg\,min}

\def \RR {{\mathbb{R}}}

\begin{document}

\twocolumn[
\icmltitle{PIDformer: Transformer Meets Control Theory}



\icmlsetsymbol{equal}{$\dagger$}

\begin{icmlauthorlist}
\icmlauthor{Tam Nguyen}{rice}
\icmlauthor{César A. Uribe}{rice}
\icmlauthor{Tan M. Nguyen}{equal,nus}
\icmlauthor{Richard G. Baraniuk}{equal,rice}
\end{icmlauthorlist}

\icmlaffiliation{rice}{Department of Electrical \& Computer Engineering, Rice University, Houston, USA}
\icmlaffiliation{nus}{Department of Mathematics, National University of Singapore, Singapore}

\icmlcorrespondingauthor{Tam Nguyen}{mn72@rice.edu}

\icmlkeywords{Machine Learning, ICML}

\vskip 0.3in
]



\printAffiliationsAndNotice{\icmlEqualContribution} 

\begin{abstract}
In this work, we address two main shortcomings of transformer architectures: input corruption and rank collapse in their output representation. We unveil self-attention as an autonomous state-space model that inherently promotes smoothness in its solutions, leading to lower-rank outputs and diminished representation capacity. Moreover, the steady-state solution of the model is sensitive to input perturbations. We incorporate a Proportional-Integral-Derivative (PID) closed-loop feedback control system with a reference point into the model to improve robustness and representation capacity. This integration aims to preserve high-frequency details while bolstering model stability, rendering it more noise-resilient. The resulting controlled state-space model is theoretically proven robust and adept at addressing the rank collapse. Motivated by this control framework, we derive a novel class of transformers, PID-controlled Transformer (PIDformer), aimed at improving robustness and mitigating the rank-collapse issue inherent in softmax transformers. We empirically evaluate the model for advantages and robustness against baseline transformers across various practical tasks, including object classification, image segmentation, and language modeling.
\end{abstract}

\vspace{-8mm}
\section{Introduction}
\label{sec:intro}
Transformer models~\cite{vaswani2017attention} have shown remarkable achievements across various domains such as reinforcement learning~\cite{chen2021decision,janner2021offline}, computer vision~\cite{dosovitskiy2021an,touvron2020deit,zhao2021point,guo2021pct}, natural language processing~\cite{devlin2018bert,al2019character,child2019generating,JMLR:v21:20-074} and other practical applications~\cite{zhang2019deep,gulati2020conformer}. At the core of transformers lies the self-attention mechanism, which computes weighted averages of token representations within a sequence based on the similarity scores between pairs of tokens, thus capturing diverse syntactic and semantic relationships effectively~\cite{cho-etal-2014-learning,parikh-etal-2016-decomposable}. This flexibility in capturing relationships has been identified as a key factor contributing to the success of transformers.

\vspace{-2mm}
\subsection{Background: Self-Attention}
\label{sec:background}
Given a sequence of tokens $\bX^{\ell}:=[\bx^{\ell}(1),\cdots,\bx^{\ell}(N)]^\top$, $\bX^\ell \in \RR^{N\times D_x}$, the query, key and value matrices at layer $\ell$-th are $\bQ^{\ell}=\bX{\bW_Q^{\ell}}^\top$; $\bK^{\ell}=\bX{\bW_K^{\ell}}^\top$; and $\bV^{\ell}=\bX{\bW_V^{\ell}}^\top$, respectively. The weight matrix $\bW^{\ell}_Q, \bW^{\ell}_K \in \RR^{D_{qk}\times D_x}$ and $\bW^{\ell}_V\in \RR^{D\times D_x}$. The attention mechanism computes the output of token $i$ at layer $\ell$-th as follows
\begin{equation}
\label{eq:attention-vec}
{{\bu}^{\ell}}(i)=\sum_{j=1}^N{\rm softmax}\Big({\bq^{\ell}}(i)^\top{\bk^{\ell}}(j)/\sqrt{D_{qk}}\Big){\bv^{\ell}}(j),
\end{equation}
where ${\bq^{\ell}}(i)$ is the row $i$-th of $\bQ^{\ell}$ and ${\bk^{\ell}}(j), {\bv^{\ell}}(j)$ are the row $j$-th of $\bK^{\ell}, \bV^{\ell}$, respectively.
The softmax function computes the attention score between token $i$ and $j$, for all $i, j = 1, \dots, N$.
The self-attention~(\ref{eq:attention-vec}) is referred to as softmax attention. Our work refers to a transformer that uses softmax attention as a softmax transformer.

Despite their remarkable success, transformers exhibit practical performance issues in their robustness and representation capacity. For example, recent studies \cite{mahmood2021robustness,madry2017towards,zhou2022understanding} have provided empirical evidence of Vision Transformer's susceptibility to adversarial attacks and common input perturbations, such as noise or blur. Additionally, deep transformer-based models have been observed to suffer from rank-collapse in their outputs, wherein token embeddings become increasingly similar as the model depth increases~\cite{shi2022revisiting, dong2021attention,wang2022antioversmoothing}. This issue severely constrains the representation capacity of transformers, hindering their performance in various tasks. Addressing these issues is crucial for ensuring the reliability and effectiveness of transformer models across different applications.

\vspace{-2mm}
\subsection{Contribution}

We introduce self-attention as a self-evolving state-space model (SSM) and provide insights into the non-robustness and rank-collapse issues inherent in transformers. Specifically, we demonstrate that self-attention can be seen as a discretization of an SSM from a gradient flow, minimizing the nonlocal total variation~\cite{Gilboa2008NonlocalOW} of an input signal and promoting smoothness. This characteristic leads to rank collapse and diminishes the output's representation capacity. Additionally, the steady-state solution of the SSM is sensitive to input perturbation. Motivated by this novel understanding, we propose the Proportional-Integral-Derivative (PID) control transformer, PIDformer, as a new transformer class that mitigates both issues. PIDformer is derived as a discretization of a PID-control integrated SSM proven to enhance the model's stability and representation capacity. Our contributions are four-fold.

\vspace{-3mm}
\begin{enumerate}
    \item  We present a novel control framework for self-attention mechanisms, unveiling the connection between self-attention and the state-space model. Our analysis sheds light on the shortcomings of transformers, which exhibit non-robust behavior to input perturbations and are prone to rank collapse. 
    \item Motivated by these analyses, we propose PIDformer, a new class of transformers, that integrates a Proportional-Integral-Derivative (PID) controller into transformers. PIDformer enhances model robustness and effectively mitigates the rank-collapse issue.
    \item We demonstrate how the connection between energy optimization and our controlled SSMs enhances the understanding of these models.
    \item We theoretically prove that employing softmax self-attention is inherently sensitive to noise and tends to produce low-rank outputs. In contrast, our controlled SSM is guaranteed to exhibit superior robustness and avoid the rank-collapse issue.
\end{enumerate}

\vspace{-2mm}
We empirically demonstrate the advantages of PIDformers on various large-scale applications, including the ImageNet object classification~\cite{deng2009imagenet} (under diverse input perturbations and robustness benchmarks), ADE20K image segmentation~\cite{zhou2018semantic}, and WikiText-103 language modeling~\cite{DBLP:conf/iclr/MerityX0S17}. tasks.

\textbf{Organization.}
We structure our paper as follows: In Section~\ref{sec:method}, we introduce a control framework for self-attention, offering insights into the non-robustness and rank-collapse issues in transformer-based models. In Section~\ref{sec:pid-control}, we incorporate a PID controller into the SSM, providing theoretical guarantees of its stability and ability to mitigate the rank-collapse issue. Subsequently, we developed PIDformer, a discretization of the PID-controlled SSM, and established the connection between these dynamics and energy optimization for further understanding. In Section~\ref{sec:experiment}, we empirically validate the benefits of PIDformer. We review related work in Section~\ref{sec:related_work}. Finally, we summarize our main contributions and provide additional results, details, and proofs in the Appendix.
\vspace{-3mm}
\section{A Control Framework for Self-Attention}
\label{sec:method}


Consider the value matrix of layer ${\ell}$-th $\bV^{\ell}:=[\bv^{\ell}(1),\cdots,\bv^{\ell}(N)]^\top \in \RR^{N \times D}$ in Section~\ref{sec:background}. Let $\Omega \subset \RR$, $x \in \Omega$, and $\bv(x, t):=[v_{1}(x, t), \dots, v_{D}(x, t)]^T$ be a real vector-valued function, $\bv:\Omega \times [0, \infty)\rightarrow \RR^{D}$, $\bv \in L^{2}(\Omega \times [0, \infty))$. Assume the value matrix $\bV^{\ell}$ discretizes the function $\bv(x, t)$ on the spatial and time dimension. In the context of a control system, $\bv(x)$ can be considered as the state signal of the following state-space model:
\begin{align}
\label{eq:state-space}
\frac{d\bv(x, t)}{dt} &=\displaystyle \int_{\Omega}(\bv(y, t) - \bv(x, t))K(x, y, t)dy + \bz(x, t) \nonumber \\
    \bv(x, 0) &= \bv^0(x), \bz(x, t) = \bm{0}, \forall x \in \Omega, \forall t \geq 0
\end{align}
where $\bm{z} \in L^{2}(\Omega \times [0, \infty))$ is a control input and $\bv^0$ is the initial state. 
The function $K(x, y, t)$ is the kernel function that captures the proximity of the signal $\bv$ at positions $x, y$ at time $t$.
Here, the SSM is autonomous, as no control inputs or feedback are fed into the system. In this section, we illustrate that system in~(\ref{eq:state-space}) induces smoothness to the signal by minimizing the nonlocal total variation~\cite{Gilboa2008NonlocalOW} of the signal, hence losing detailed information as it evolves. Subsequently, we show that self-attention serves as a discretization of this dynamic. Lastly, we theoretically demonstrate that the SSM in~\ref{eq:state-space} is vulnerable to input perturbation and representation collapse.

\vspace{-2mm}
\subsection{Connection between State Space Model and Nonlocal Variational Minimization}
\label{subsec:smooth-state-space}

We show that the gradient flow aimed at minimizing the following nonlocal functional is a case of our SSM described in~(\ref{eq:state-space})
\begin{equation}
    J(\bv) = \frac{1}{2} \int_{\Omega \times \Omega} \|\bv(x) - \bv(y)\|_2^2 k(x, y)dxdy.
\end{equation}
Here, $J(\bv)$, the sum of the square of the nonlocal derivative on the spatial dimension $\partial_y\bv(x) = \bigl(\bv(x) - \bv(y)\bigl)\sqrt{k(x,y)}$~\cite{Gilboa2008NonlocalOW}
, represents the non-local variation of the signal $\bv$. $k(x, y)$ captures the proximity between position $x$ and $y$ in the signal. Minimizing $J(\bv)$ promotes the smoothness of $\bv$ and penalizes high-frequency in the signal.

The gradient of $J$ with respect to $\bv$ is then given by
\begin{equation}
\begin{aligned}
\label{eqn:devJu}
    \nabla_{\bv} J(\bv) = \left[\frac{\partial J}{\partial v_1}, \frac{\partial J}{\partial v_2},  \dots, \frac{\partial J}{\partial v_{D}} \right]^T.
\end{aligned}
\end{equation}
As shown in the Appendix~\ref{secapp:djv}, the Frechet derivative of $J$ with respect to $v_j$ is
\begin{align}  
\label{eq:partial-dev}
\frac{\partial J}{\partial v_j} 
&= \int_{\Omega}(v_j(x) - v_j(y)(k(x, y) + k(y, x))dy. 
\end{align}
Substituting the formula for ${\partial J}/{\partial v_j}$ in~(\ref{eq:partial-dev}) into~(\ref{eqn:devJu}) for $\nabla_{\bv} J(\bv)(x)$, we obtain the following gradient flow
\begin{equation}
\begin{aligned}
    \label{eq:gradient-descent}
    \frac{d\bv(x,t)}{dt} &= -\nabla_{\bm{v}}J(\bv) \\
    &= \int_{\Omega} \bigl(\bm{v}(y,t) - \bm{v}(x,t) \bigl)\bigl(k(x, y) + k(y,x)\bigl)dy,
\end{aligned}
\end{equation}
The autonomous state-space representation in~(\ref{eq:state-space}) simplifies to this dynamic when $K(x, y, t) := k(x, y) + k(y, x)$, which is symmetric and time-invariant. In this scenario, the model reduces the total nonlocal variance of the signal, resulting in a smoother solution. This renders the model susceptible to rank collapse in the output representation. In Section~\ref{subsec:state-space-analysis}, we prove that the model suffers from rank collapse regardless of whether $K(x, y, t)$ is symmetric.

\textbf{Connection between SSM and self-attention.} We show that a discretization of our SSM recovers the self-attention mechanism. Let $\bq, \bk:\Omega \times [0, \infty) \rightarrow \RR^{D_{qk}}$, $\bq,\bk \in L^{2}(\Omega \times [0, \infty))$ be real vector-valued functions. Similar to $\bv(x, t)$, we can discretize $\bq(x, t),\bk(x, t)$ on spatial dimension to attain the query vectors $\bq^{\ell}(1), \dots, \bq^{\ell}(N) \in \RR^{D_{qk}}$, and the key vectors $\bk^{\ell}(1), \dots, \bk^{\ell}(N) \in \RR^{D_{qk}}$ of layer $\ell$-th. We define the proximity kernel as
\[
    K(x,y,t):= \frac{\exp\bigl(\bq(x, t)^{T}\bk(y, t)/\sqrt{D_{qk}}\bigl)}{\int_{\Omega}\exp\bigl(\bq(x, t)^{T}\bk(y', t)/\sqrt{D_{qk}}\bigl)dy'}.
\]
Applying the Euler method to discretize~(\ref{eq:state-space}) with the time step $\Delta t(x):= 1$, the update step of the system becomes
\begin{equation}
   \begin{aligned}
\label{eqn:sym-attn-cont}
&\bm{v}(x, t + 1) \\
\approx
&\displaystyle \int_{\Omega}\frac{\exp\bigl(\bq(x, t)^{T}\bk(y, t)/\sqrt{D_{qk}}\bigl)}{\int_{\Omega}\exp\bigl(\bq(x, t)^{T}\bk(y', t)/\sqrt{D_{qk}}\bigl)dy'} \bv(y, t)dy.
\end{aligned} 
\end{equation}

Using the Monte-Carlo method~\cite{13ab5b5e-0237-33fb-a7a8-6f6e4e0d4e0f} to approximate the integrals in the spatial dimension in~(\ref{eqn:sym-attn-cont}), we attain 
\begin{equation}
\nonumber
\bm{v}^{\ell + 1}(i) \approx \sum_{j=1}^N{\rm softmax}\Big({\bq}^{\ell}(i)^\top{\bk}^{\ell}(j)/\sqrt{D_{qk}}\Big){\bv}^{\ell}(j).
\end{equation}
which recovers $\bu^{\ell}(i)$, the output token $i$ of self-attention at layer $\ell$-th as in~(\ref{eq:attention-vec}). As self-attention discretizes the SSM outlined in~(\ref{eq:state-space}), it inherits the characteristics of the model, making it susceptible to input corruption and output rank collapse. These properties are theoretically demonstrated in Section~\ref{subsec:state-space-analysis}.

\vspace{-2mm}
\subsection{Stability and Representation Collapse of the State Space Model}
Model robustness is its ability to maintain high performance despite encountering uncertain or challenging scenarios such as noisy data, distribution shifts, or adversarial attacks~\cite{wang-bansal-2018-robust,dong2020benchmarking}.
Robustness also entails stability, wherein the model's output remains relatively unchanged even when the input is perturbed.

\label{subsec:state-space-analysis}
For the theoretical analysis of our SSMs, we assume that the kernel $K$ is time-invariant, i.e.,
$K(x,y,t)=K(x,y)$. This assumption is practical in the context of transformers, particularly in deep transformer models, where the attention matrix tends to remain similar after the initial layers~\cite{shi2022revisiting}.
The discretization of model in~(\ref{eq:state-space}) on the spatial dimension gives
\begin{equation}
\nonumber
    \displaystyle \frac{d\bv(i, t)}{dt} =\displaystyle \sum_{j = 1}^N(\bv(j, t) - \bv(i, t))K(i, j), 
\end{equation}
for $i,j = 1, 2, \dots, N$
By choosing $K(i,j):= {\rm softmax}\bigl(\bq(i)^{T}\bk(j)/\sqrt{D_{qk}}\bigl)$, its corresponding matrix representation is obtained as
\begin{align}
\label{eq:ode1}
    \bV'(t){dt} = \bK\bV(t) - \bV(t), \bV(0) = \bV^0,
\end{align}
where $\bm{K}$ is a right-stochastic matrix with all positive entries. In the context of transformer, $\bm{K}$ is the attention matrix and $\bV = [\bv^0(1), \dots, \bv^0(N)]^T$ is the value matrix at the first layer.
Lemma~\ref{lem:state-space-sol} sheds light on the stability and representation collapse of the solution for the SSM in~(\ref{eq:state-space}).
\begin{lemma}
    \label{lem:state-space-sol}
    Given $\{\alpha_1, \alpha_2,\dots, \alpha_M\}, M \leq N$, is the complex spectrum of $\bK - \bI \in \mathbb{R}^{N \times N}$. The solution of the ordinary differential equation (ODE)~(\ref{eq:ode1}) is given by
    \begin{align}
    \label{eq:state-space-sol}
        \bV(t) = \bm{P}\mathrm{exp}(\bm{J}t)\bm{P}^{-1}\bV^0,
    \end{align}
where $\bm{P}\bm{J}\bm{P}^{-1}$ is the Jordan decomposition of $\bm{K} - \bm{I}$, $\bm{P}$ is invertible and contains the generalized eigenvectors of $\bm{K} - \bm{I}$, and $\bm{J} = \bm{\mathrm{diag}}(\bm{J}_{\alpha_1, m_1}, \bm{J}_{\alpha_2, m_2}, \dots, \bm{J}_{\alpha_M, m_M}) $ is the Jordan form of matrix $\bm{K} - \bm{I}$ with,\\
$\bm{J}_{\alpha_i, m_i} = \begin{bmatrix} 
    \alpha_i &  1& \dots &0\\
   \vdots & \ddots & & \vdots\\
    &  & \alpha_i& 1\\
    0 & \dots    &   & \alpha_i
    \end{bmatrix} \in \mathbb{R}^{m_i \times m_i}$, for $i = 1, \dots, M$ are Jordan blocks. Here, $\sum_{i = 1}^Mm_i = N$.
\end{lemma}

The proof of Lemma~\ref{lem:state-space-sol} is shown in the Appendix~\ref{secapp:state-space-sol}. Since $\bK$ is a positive right-stochastic matrix, its largest and unique eigenvalue $\alpha_1$ is 1 and $|\alpha_i| < 1$ (see Theorem 4.1 in~\cite{strohmer2020mathdl}), meaning $\mathrm{Re(\alpha_i)} \in [-1, 1)$, for $i = 2, \dots, M$. Hence, the matrix $\bK - \bI$, whose eigenvalues are $\alpha_1 - 1, \dots, \alpha_M - 1$, has a unique largest eigenvalue of 0 and the real part of other eigenvalues in $[-2, 0)$. This leads to the rank collapse of the steady-state solution, as stated in the following Lemma~\ref{lem:steady-state-state-space-sol}.
\begin{lemma}
\label{lem:steady-state-state-space-sol}
     $\lim_{t \to \infty}\bV(t) = \begin{bmatrix} \displaystyle c_{1,1}\bm{p_1}, & \dots,&c_{1, D_x}\bm{p_1} \end{bmatrix}$,
    where $\bm{p}_1$ is the eigenvector corresponds with the eigenvalue $(\alpha_1 - 1) = 0$ of $\bm{K} - \bm{I}$, and $c_{1,1}, \dots, c_{1,D_x}$ are the coefficients w.r.t $\bm{p}_1$ of the decomposition of $\bm{V}^0$'s columns in the Jordan basis (column vectors of $\bm{P}$).
\end{lemma}

\vspace{-2mm}
The proof of Lemma~\ref{lem:steady-state-state-space-sol} is shown in the Appendix~\ref{secapp:steady-state-state-space-sol}.
This yields two insights. Firstly, the steady-state solution of the system depends on the initial $\bm{V}^0$, implying that any perturbation in the input results in changes in the output. 
Secondly, the solution experiences rank collapse, with the rank of its steady state solution being 1 as $t \to \infty$. This indicates that our SSM in (\ref{eq:state-space}) not only yields a non-robust solution but also experiences information loss (low-rank output representation). As the self-attention mechanism discretizes the model in (\ref{eq:state-space}), it inherently exhibits both issues.
\vspace{-3mm}
\section{Transformer with PID-Controller for State-Space Representation}
\label{sec:pid-control}
To counteract the loss of detailed information caused by smoothness and to bolster model stability, a PID controller is integrated into the state-space representation as follows:
\begin{align}
\label{eq:state-space-pid}
    \displaystyle \frac{d\bv(x, t)}{dt} &=\displaystyle \int_{\Omega}(\bv(y, t) - \bv(x, t))K(x, y, t)dy + \bm{z}(x, t) \nonumber \\
    \bm{z}(x, t) &=\displaystyle \lambda_P\bm{e}(x, t) + \lambda_I\int_0^{t}\bm{e}(x, t) + \lambda_D\frac{d\bm{e}(x, t)}{dt} \nonumber \\
    \bv(x, 0) &= \bv^0(x), \bz(x, 0) = \bm{0}.
\end{align}

The regularizer term, denoted as $\bm{e}(x, t) = \bff(x) - \bv(x, t)$, encapsulates the loss of information as $\bv(x, t)$ becomes smoother over time. Here, the reference function $\bff(x)$ represents a high-frequency signal containing detailed information about the original inputs. We select $\bff(x)$ as the scaled initial value function, denoted as $\beta\bv(x, 0)$. In the context of a transformer, we set $\bff(i) = \beta\bv^0(i)$, representing the value vector embedding at token index $i$ of the first layer. This choice is motivated by our desire to have flexibility in determining the detailed information from the input signal we wish to preserve. This flexibility is governed by the parameter $\beta \in (0, 1]$. The regularizer $\bm{e}(x, t)$ is fed back into the system, guiding the model to reintegrate the lost information while maintaining stability through three components: (P), (I), and (D).

\vspace{-3mm}
\begin{itemize}
    \item The (P) term is directly proportional to the regularizer, $e(x, t)$. In cases of substantial information loss, the control input $\bz(x, t)$ should be proportionately large, determined by the gain factor $\lambda_P$, to reintroduce the lost information into the system. A small choice of $\lambda_P$ results in slow convergence, while a large choice may lead to overshooting issues, causing instability in reaching the reference point.
    \item The (I) term accumulates all past errors, given by $\lambda_I\int_0^{t}\bm{e}(x, t)$. This component aids in reintroducing any persistent, long-term loss of information that might persist despite proportional control.
    \item Finally, the (D) term, $\displaystyle \lambda_D\frac{d\bm{e}(x, t)}{dt}$, anticipates future losses of information by considering the rate at which the error is changing. A more rapid change in error prompts a greater control effect, and the derivative term proves beneficial in enhancing the stability and responsiveness of the control system.
\end{itemize}

\vspace{-3mm}
In this section, we unveil a connection between the two components, (P) and (I), of the SSM in~(\ref{eq:state-space-pid}) and different optimization methods applied to minimize a regularized functional. This functional is tailored to preserve the detailed information of the solution. Moreover, we show that the P-control (where $\lambda_I = \lambda_D = 0$), PD-control ($\lambda_I = 0$), and PID-controlled SSM in~(\ref{eq:state-space-pid}) are theoretically guaranteed to be more robust and mitigate the issue of rank collapse. Subsequently, we introduce the PID-controlled transformer (PIDformer), a novel architecture that enhances performance and robustness.

\vspace{-3mm}
\subsection{Connection between (P) and (I) Components with Different Optimization Methods} 

In Section~\ref{subsec:smooth-state-space}, we have shown that the SSM in~(\ref{eq:state-space}) implicitly performs a gradient descent to minimize the nonlocal variation $J(\bv)$, which leads to the loss of signal information. Now, we illustrate that the feedback-controlled state-space in~(\ref{eq:state-space-pid}), without the derivative (D) ($\lambda_D = 0$), implicitly minimizes the following functional:
\begin{equation}
  \begin{aligned}
    \label{eqn:energy-functional}
    E(\bv, \bff) &= J(\bv) + G(\bv,\bff) \\ 
    &= \frac{1}{2} \int_{\Omega \times \Omega} \|\bv(x) - \bv(y)\|_2^2 k(x, y)dx dy \\
    &+ \frac{\lambda}{2}  \int_{\Omega} \|\bv(x) - \bff(x)\|_2^2 dx.
    \end{aligned}  
\end{equation}
where the data fidelity term $G(\bv,\bff) = \frac{\lambda}{2} \int_{\Omega} \|\bv(x) - \bff(x)\|_2^2 dx$~\cite{Gilboa2008NonlocalOW, Gilboa2007NonlocalLI} is introduced to penalize significant information loss. This observation further validates that systems in~(\ref{eq:state-space-pid}) retains relevant information from the reference signal $\bff$.

\textbf{P-controlled SSM as gradient descent to minimize $E(\bv, \bff)$}.
The gradient of $E$ w.r.t $\bv$ is $\nabla_{\bv}E(\bv) = \nabla_{\bv}J(\bv) + \lambda \bigl(\bv(x) - \bff(x)\bigl)$.
The derivation of the derivative is given in Appendix~\ref{secapp:derE}.
Using the gradient descent method, we obtain the gradient flow:
\\
\begin{equation}
\label{eq:flowE}
    \begin{aligned}
        &\frac{d\bv(x,t)}{dt} = -\nabla_{\bm{u}}E(\bv) \\
        &= \int_{\Omega} \bigl(\bm{v}(y,t) - \bm{v}(x,t) \bigl)\bigl(k(x, y) + k(y,x)\bigl)dy \\
        &+ \lambda\bigl(\bff(x) - \bv(x, t)\bigl).
    \end{aligned}
\end{equation}
If we set
$K(x, y, t) := k(x, y) + k(y, x)$ to be symmetric and time-invariant, and $\lambda_P = \lambda, \lambda_I = \lambda_D = 0$, the controlled system in~(\ref{eq:state-space-pid}) simplifies to the gradient flow of $E$ in~(\ref{eq:flowE}). It suggests that integrating the (P) component into the system in~(\ref{eq:state-space}) minimizes the functional $E$ and reintroduces the lost information to the system.

\textbf{PI-controlled SSM as Bregman iteration to minimize $E(\bv, \bff)$}.
An alternative to gradient descent, Bregman iteration ~\cite{stanbregman,zhangnonlocalbregman} iteratively refines the solution by minimizing a Bregman divergence, which measures the discrepancy between the current solution and a reference point.
Given the convex functional $J(\bv)$, the Bregman divergence of $J$ between $\bv$ and $\bm{s} \in L^{2}(\Omega)$ is $D_J^p(\bv, \bm{s}) := J(\bv) - J(\bm{s}) - 
\langle  \bm{p} , \bv - \bm{s} \rangle $, where $\bm{p} \in \partial J(\bm{s})$, the subgradient of $J$ at $\bm{s}$. $D_J^p(\bv, \bm{s})$ captures the difference between $J(\bv)$ and the tangent plane $J(\bm{s}) - \langle\bm{p} , \bv - \bm{s}\rangle$.
The $\ell + 1$-th Bregman iteration to minimize $\min_{\bv}J(\bv)$ with the contraint $G(\bv, \bff)$ is given by:
\begin{align}\label{eqn:breg-op}
    \bv^{\ell + 1} &{=} \argmin_{\bv} D_J^{p^{\ell}}(\bv, \bv^{\ell}) + G(\bv, \bff), p^{\ell} \in \partial J(\bv^{\ell})
\end{align}

The following Lemma~\ref{lem:breg-iter} shows that the optimization problem in~(\ref{eqn:breg-op}) can be turned into solving iterative subproblems.
\begin{lemma}
\label{lem:breg-iter}
Applying Bregman iteration to minimize $E(\bv, \bff)$ involves solving iterative subproblems:
\begin{align} 
    \bv^{\ell + 1} &= \displaystyle \argmin_{\bv} J(\bv) + \frac{\lambda}{2} \int_{\Omega} \|\bv(x) - \bff(x) - \bm{e}^{\ell}_{\mathrm{a}}(x)\|_2^2 dx \nonumber \\
    \bm{e}^{\ell}_{\mathrm{a}}(x) &= \sum_{m = 1}^{\ell} \bm{e}^m(x) = \sum_{m = 1}^{\ell}\big( \bff(x) - \bv^m(x) \big),
 \end{align}
\end{lemma}
The proof of Lemma~\ref{lem:breg-iter} is in Appendix~\ref{secapp:breg-iter}.
Here, the term $\bm{e}^{\ell}_{\mathrm{a}}(x)$ captures the accumulated information loss between the original and the smoothed signals $\bv^m(x)$ of each iteration $m = 1, \dots, \ell$.
Taking a one-step update in the direction of gradient descent (see Appendix~\ref{secapp:grad-flow-breg}), we obtain
\begin{align}
   \label{eq:dis_breg}
   \bv^{\ell + 1}(x) 
   &= \int_{\Omega} \bigl(\bm{v}^{\ell}(y) - \bm{v}^{\ell}(x) \bigl)\bigl(k(x, y) + k(y,x)\bigl)dy \nonumber \\
   & + \bv^{\ell}(x) + \lambda \bm{e}^{\ell}(x) + \lambda\bm{e}^{\ell}_{\mathrm{a}}(x).
\end{align}
On the other hand, the Euler discretization with $\Delta t = 1$ of the PI-controlled state-space in~(\ref{eq:state-space-pid}) (as $\lambda_D = 0$) is:
\begin{equation}
\begin{aligned}
\label{eq:dis-PI}
    \bv^{\ell + 1}(x) &= \bv^{\ell}(x) + \int_{\Omega} \bigl(\bm{v}^{\ell}(y) - \bm{v}^{\ell}(x) \bigl)K(x, y)dy \\
    &+ \lambda_P \bm{e}^{\ell}(x) + \lambda_I \sum_{m = 1}^{\ell} \bm{e}^m(x).
\end{aligned}
\end{equation}
By selecting a time-independent $K(x, y, t) := k(x, y) + k(y, x)$ and setting $\lambda_P = \lambda_I = \lambda$, the update step of the PI-controlled model in~(\ref{eq:dis-PI}) simplifies to the update step of Bregman iteration in~(\ref{eq:dis_breg}). This connection suggests that the PI-controlled SSM minimizes $E(\bv, \bff)$.
\subsection{Stability and Representation Collapse of PID-Controlled State Space Model}

In this section, we aim to show that: (i) Integrating the (P) term enhances robustness against input perturbations and mitigates rank collapse of the output representation; (ii) Adding the (D) term in PD-control further stabilizes the system by mitigating rapid and unstable changes of $\bV(t)$, (iii) finally, integrating the (I) term in the PID-controlled SSM described in~(\ref{eq:state-space-pid}) guarantees system stability, making it robust to input corruption. Following the same assumption in Section~\ref{subsec:state-space-analysis}, we assume that $K(x, y, t)$ is time-invariant for our theoretical analysis in this section.
\subsubsection{Analysis of P-control SSM}
\label{subsec:PD-robust}
\textbf{Robustness of P-controlled SSM.}
From the SSM in~(\ref{eq:state-space-pid}), by choosing $\lambda_I = \lambda_D = 0$, and applying Euler discretization on the spatial domain, the P-control model is given as:
\begin{equation}
   \begin{aligned}
\label{eq:state-space-P}
    \displaystyle \frac{d\bv(i, t)}{dt} &=\displaystyle \sum_{j = 1}^N(\bv(j, t) - \bv(i, t))K(i, j) \\
    &+ \lambda_P \bigl(\bff(i) -  \bv(i, t)\bigl), 
\end{aligned} 
\end{equation}
for $i,j = 1, 2, \dots, N$, and $K(i,j):= {\rm softmax}\bigl(\bq(i)^{T}\bk(j)/\sqrt{D_{qk}}\bigl)$. The corresponding matrix representation is given as
\begin{equation}
\begin{aligned}
    \label{eq:odeP}
    \frac{d\bV(t)}{dt} = \bK\bV(t) - (\lambda_P + 1)\bV(t) + \lambda_P \bm{F}, \bV(0) = \bV^0.
\end{aligned}
\end{equation}
where $\bm{F} = [ \bff(1), \dots, \bff(N) ]^T$.
The following Lemma~\ref{lem:state-space-P-sol} help us analyze the stability and representation collapse of the solution for the SSM in~(\ref{eq:odeP}).
Here, since the eigenvalues of $\bK$ the has the real part in $[0, 1]$, $\lambda_P + 1$ ($\lambda_P > 0$) can not be one of them. This implies that $\mathrm{det}(K - (\lambda_P + 1)I) \neq 0$ hence the matrix is non-singular.
\begin{lemma}
    \label{lem:state-space-P-sol}
    Let $\bm{B} := \bK - (\lambda_P + 1)\bI \in \mathbb{R}^{N \times N}$, the solution of the ordinary differential equation~(\ref{eq:odeP}) is
    \begin{align}
    \label{eq:state-space-P-sol}
    \bV(t) = \mathrm{exp}(\bm{B}t)(\bV^0 + \bm{B}^{-1}\bm{F}) - \lambda_P\bm{B}^{-1}\bm{F}.
    \end{align} If $\bm{B}$ has only eigenvalues with negative real parts, then $\lim_{t \to \infty}\bm{V}(t) = -\lambda_P\bm{B}^{-1}\bm{F}$.
\end{lemma}
The proof of Lemma~\ref{lem:state-space-P-sol} is shown in the Appendix~\ref{secapp:state-space-P-sol}. As shown in Section~\ref{subsec:state-space-analysis}, since the eigenvalues of $\bm{K}$ has $\mathrm{Re}(\alpha_i) \in [-1, 1], i = 1, \dots, M$, therefore the real parts of eigenvalues of $\bm{B}$ must be in the range $[-2 - \lambda_P, -\lambda_p]$, which are all negative. As the result of~\ref{lem:state-space-P-sol}, the steady state solution in~(\ref{eq:state-space-P-sol}) $\lim_{t \to \infty}\bm{V}(t) = -\lambda_P\bm{B}^{-1}\bm{F}$. Therefore, adding any perturbation to the initial state $\bV^0$ does not change the steady state solution. However, in our context of a transformer, the perturbation also affects the reference point $\bm{F}$, which is chosen to be a scaled $\beta\bm{\bV}^0$, leading to the steady state solution becomes $-\lambda_P\beta\bm{B}^{-1}\bm{\bV}^0$. Fortunately, the P-control system allows the error caused by perturbation to be as neglectable as desired. The following Proposition~\ref{lem:lambda_P} confirms the robustness of the P-control SSM.
\begin{proposition}
\label{lem:lambda_P}
    Given the coefficient $\lambda_P > 0$ in~(\ref{eq:state-space-pid}), and any arbitrary $\Bar{\epsilon}, \delta > 0$, by adding  the perturbation $\bm{\epsilon} \in \mathbb{R}^{N \times D}, \| \bm{\epsilon} \|_{\infty} \leq \Bar{\epsilon}$ to $\bV^0$, the corresponding change in the steady state solution of the system in~(\ref{eq:odeP}) is independent of $\lambda_P$ and becomes negligible with an amount of at most $\delta$ if 
    \begin{align}
        \beta \leq {\delta}/{\bar{\epsilon}}.
    \end{align}
\end{proposition}
\vspace{-2mm}
The proof of Proposition~\ref{lem:lambda_P} is shown in the Appendix~\ref{secapp:lambda_P}.
Proposition~\ref{lem:lambda_P} suggests that we can select the hyper-parameter $\beta$ to make the impact of input perturbation on the output as small as desired.

\textbf{P-controlled SSM on representation collapse.} Since $\bm{B}^{-1}$ is full rank ($\bm{B}$ is non-singular), hence $\mathrm{rank}(-\lambda_P\bm{B}^{-1}\bm{F}) = \mathrm{rank}(\bm{F})$~\cite{strang2006linear}. In the case of a transformer, when choosing $\bm{F} = \beta \bV^0$, the rank of the steady state solution equals the rank of the input $\bV^0$. This implies that the P-control dynamic in~(\ref{eq:odeP}) prevents rank collapse.
\subsubsection{Analysis of PD-controlled SSM}
Since $\lambda_D \frac{d\bm{e}(x, t)}{dt} = \lambda_D \frac{d}{dt}(\bm{f}(x) - \bm{v}(x, t)) = -\lambda_D\frac{d\bm{v}(x, t)}{dt}$,
from the SSM in~(\ref{eq:state-space-pid}), by choosing $\lambda_I = 0$, $K(i,j):= {\rm softmax}\bigl(\bq(i)^{T}\bk(j)/\sqrt{D_{qk}}\bigl)$ for $i,j = 1, 2, \dots, N$, and applying Euler discretization on the spatial domain, the PD-control model can be represented in the matrix form:
\begin{equation}
\label{eq:odePD}
    \begin{aligned}
    \bV'(t) &= \bK\bV(t) - (\lambda_P + 1)\bV(t) + \lambda_P \bm{F} - \lambda_D\bV'(t)\\
    &= \frac{1}{1 + \lambda_D}\bigl(\bm{K} - (\lambda_P + 1)\bm{I}\bigl)\bV(t) + \frac{\lambda_P}{1 + \lambda_D}\bm{F},
\end{aligned}
\end{equation}
with $\bV(0) = \bV^0$.
The solution of~(\ref{eq:odePD}) is provided in the following Lemma~\ref{lem:state-space-PD-sol}.
\begin{lemma}
    \label{lem:state-space-PD-sol}
    Let $\bm{B} := \bK - (\lambda_P + 1)\bI \in \mathbb{R}^{N \times N}$, the solution of the ordinary differential equation~(\ref{eq:odePD}) is 
    \begin{align}
    \nonumber
    \label{eq:state-space-PD-sol}
    \bV(t) = \mathrm{exp}(\frac{1}{1 + \lambda_D}\bm{B}t)(\bV^0 + \bm{B}^{-1}\bm{F}) - \lambda_P\bm{B}^{-1}\bm{F}.
    \end{align} 
and $\lim_{t \to \infty}\bm{V}(t) = -\lambda_P\bm{B}^{-1}\bm{F}$.
\end{lemma}
The proof of Lemma~\ref{lem:state-space-PD-sol} is provided in Appendix~\ref{secapp:state-space-PD-sol}. This intriguing result suggests two key insights. Firstly, incorporating the (D) component into the P-control system does not alter the steady state of the solution. Consequently, the solution of the PD-controlled SSM retains the advantages of a P-control model, including avoiding rank collapse and ensuring bounded error. Secondly, the derivative term offers an additional benefit of further stabilizing the system by decreasing the eigenvalue by a factor of $\displaystyle{1}/({1 + \lambda_D})$, thereby mitigating rapid changes in $\bV(t)$.
\subsubsection{Analysis of PID-controlled SSM}

Following the same analysis in Section~\ref{subsec:PD-robust}, by choosing $K(i,j):= {\rm softmax}\bigl(\bq(i)^{T}\bk(j)/\sqrt{D_{qk}}\bigl)$ and discretizing on the spatial domain, the matrix representation of the PID-controlled SSM reduced from~(\ref{eq:state-space-pid}) becomes
\begin{equation}
 \begin{aligned}
\label{eq:matrix-PI}
   \bV'(t) =&\frac{1}{\lambda_D + 1} \Bigg( \bigl(\bK - (\lambda_P + 1)\bm{I}\bigl)\bV(t) \\
   &+ \lambda_I \int_{0}^t (\bm{F} - \bV(t))dt
    + \lambda_P \bm{F} \Bigg),
\end{aligned}   
\end{equation}
where $\bV(0) = \bV^0$. To deal with the integral in~(\ref{eq:matrix-PI}), we take the derivative of both sides, the equation becomes $\bV''(t) = \displaystyle \frac{1}{\lambda_D + 1} \big(\big(\bK - (\lambda_P + 1)I \big)\bV'(t) - \lambda_I\bV(t) \big)$, which is turned into a system of 1-st order differential equation:
\begin{equation}
\begin{aligned}
    \label{eq:ode-pi}
    \begin{bmatrix} \bV'(t) \\ \bV''(t) \end{bmatrix} &= \begin{bmatrix} \bm{0} & \bm{I} \\ \displaystyle -\frac{\lambda_I\bm{I}}{\lambda_D + 1} &\displaystyle \frac{\bm{K} - (\lambda_P + 1)\bm{I}}{\lambda_D + 1} \end{bmatrix} \begin{bmatrix} \bV(t) \\ \bV'(t) \end{bmatrix},
\end{aligned}
\end{equation}
where $\bV(0) = \bV^0$, and $\bV'(0) =\displaystyle \frac{1}{\lambda_D + 1} \big( (\bm{K} - (\lambda_P + 1))\bV^0 + \lambda_P\bm{F} \big)$.
To gain robustness, the steady state solution of the model should be independent of any perturbation of the input $\bV_0$
The following Proposition~\ref{lem:lambda_PI} illustrates the stability of the system.
\begin{proposition}
    \label{lem:lambda_PI} For any $\lambda_P, 
    \lambda_I, \lambda_D > 0$, the system in~(\ref{eq:ode-pi}) has a stable solution.
\end{proposition}
The proof of Proposition~\ref{lem:lambda_PI} is in the Appendix~\ref{secapp:lambda_PI}. 
The Proposition implies that the PID-controlled SSM in~(\ref{eq:state-space-pid}) remains robust and stable for any selection of positive values for $\lambda_P, \lambda_I, \lambda_D$.

\subsection{Transformer with PID Control}
\begin{figure}[!t]
\centering
\includegraphics[width=0.48\textwidth]{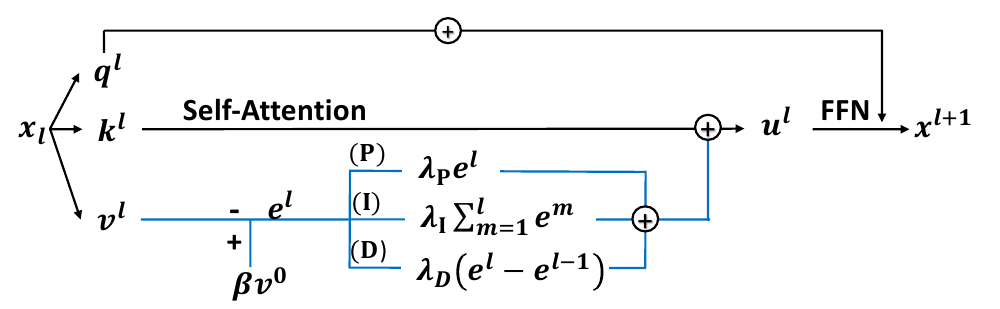}
\vspace{-0.3in}
\caption{\small Our proposed PIDformer model at each layer.}
\label{fig:pid-illustrate} 
\end{figure}

By applying the Euler discretization with time step $\Delta t = 1$, initializing $\bv$ at $t = 0$ as $\bv(x, 0) = v^0(x)$, and choosing 
\[
K(x,y, t):= \frac{\exp\bigl(\bq(x, t)^{T}\bk(y, t)/\sqrt{D_{qk}}\bigl)}{\int_{\Omega}\exp\bigl(\bq(x, t)^{T}\bk(y', t)/\sqrt{D_{qk}}\bigl)dy'},
\]
the update step of PID-controlled SSM in~(\ref{eq:state-space-pid}) becomes:
\begin{equation}
  \begin{aligned}
\label{eq:pid-dis-t}
    &\bv^{\ell + 1}(x) \\
    &\approx  \int_{\Omega} \bigl(\bm{v}^{\ell}(y) - \bm{v}^{\ell}(x) \bigl)\displaystyle\frac{\exp\bigl(\bq^{\ell}(x)^{T}\bk^{\ell}(y)/\sqrt{D_{qk}}\bigl)}{\int_{\Omega}\exp\bigl(\bq^{\ell}(x)^{T}\bk^{\ell}(y')/\sqrt{D_{qk}}\bigl)dy'}dy \\
    &+ \bv^{\ell}(x) + \lambda_P \bm{e}^{\ell}(x) + \lambda_I \sum_{m = 1}^{\ell} \bm{e}^m(x) + \lambda_D(\bm{e}^{\ell}(x) - \bm{e}^{\ell - 1}(x)),
\end{aligned}  
\end{equation}
where $\bm{e}^m(x) = \bff(x) - \bv^m(x)$ for $m = 1,\dots, \ell$. 
Applying the Monte-Carlo method to approximate the integrals in~(\ref{eq:pid-dis-t}) and discretizing $\bv^{l + 1}(x)$, $\bv^m(x)$, and $\bv_0(x)$ on a spatial dimension, and by choosing $\bff(x) = \bv(x)$, we attain the output of the following novel PID-attention at layer $\ell$-th is defined as
\begin{definition}[PID-control Transformer (PIDformer)]
\label{def:pid-attn}
Given a set of key and value vectors $\{\bk^{\ell}(j),\bv^{\ell}(j)\}_{j=1}^{N}$ in each layer $\ell$, $\ell=1,\dots,L$, for each query vector $\bq^{\ell}(i)$, $i=1,\dots,N$, in the same layer, the self-attention unit at layer $\ell$ in a PID-control Transformer (PIDformer) computes the corresponding output vector $\bu^{\ell}(i)$ of the query $\bq^{\ell}(i)$ by the following attention formula:
\begin{equation}
   \begin{aligned}
    \bu^{\ell}(i)  
    &=\sum_{j = 1}^N {\rm softmax}\Big({\bq}^{\ell}(i)^\top{\bk}^{\ell}(j)/\sqrt{D_{qk}}\Big)\bm{v}^{\ell}(y) \\
    &+ \lambda_P \bm{e}^{\ell}(i) + \lambda_I \sum_{m = 1}^{\ell} \bm{e}^{m}(i) + \lambda_D(\bm{e}^{\ell}(i) - \bm{e}^{\ell - 1}(i)),
\end{aligned} 
\end{equation}
where $\bm{e}^{\ell} = \bv^0 - \bv^{\ell}$, $\bv^{0}(1),\dots, \bv^{0}(N)\in \RR^{D}$ are the value vectors in the first layer of PIDformer. 
\end{definition}
Since PID-attention is a discretization of the controlled SSM in (\ref{eq:state-space-pid}), it is inherently a more robust attention mechanism.
Fig.~\ref{fig:pid-illustrate} illustrates the architecture of PIDformer.

\vspace{-3mm}
\section{Experimental Results}
\label{sec:experiment}
In this section, we empirically demonstrate the advantages of our proposed PIDformer approach across multiple tasks, including ImageNet classification~\cite{deng2009imagenet}, ADE20K image segmentation~\cite{zhou2018semantic}, and language modeling on WikiText-103~\cite{DBLP:conf/iclr/MerityX0S17}. Our objectives are to: (i) illustrate that PIDformer significantly outperforms the transformer baseline with softmax-attention across diverse tasks, (ii) highlight that the PID DeiT model exhibits significantly higher robustness than softmax attention under various adversarial attacks, and for out-of-distribution generalization, (iii) demonstrate that PID DeiT does not suffer from rank collapse in output representation.
Throughout our experiments, we compare the performance of our proposed models with baselines of the same configuration. For additional details regarding datasets, models, and training procedures, please refer to Appendix~\ref{secapp:experiments}.

\textbf{Object Classification on ImageNet.}
To demonstrate the advantage of our PIDformer, we compare it with the DeiT baseline~\cite{touvron2020deit} on the ImageNet image classification task. Our PID DeiT surpasses the DeiT baseline, as shown in Table~\ref{tab:attacks}. Notably, our model performs significantly better than the baseline under white-box attacks, including fast gradient sign method (FGSM)~\cite{dong2020benchmarking}, projected gradient descent method (PGD)~\cite{tramer2019adversarial}; score-based black-box attack method SPSA~\cite{pmlr-v80-uesato18a}; and sparse $L_1$ descent (SLD)~\cite{NEURIPS2019_5d4ae76f} method as well as noise-adding attack. Moreover, the last four rows of Table~\ref{tab:attacks} demonstrate that PID DeiT is consistently more robust than the DeiT baseline under other adversarial examples and out-of-distribution dataset, including the Imagenet-C (common data corruption and perturbations, such as adding noise and blurring the images)~\cite{hendrycks2019benchmarking}, Imagenet-A (adversarial examples)~\cite{hendrycks2021natural}, Imagenet-R (out of distribution generalization)~\cite{hendrycks2021many}, and Imagenet-O(out-of-distribution detection)~\cite{hendrycks2021natural} datasets. Furthermore, in Appendix~\ref{secapp:moreattacks}, we visualize the performance gap between PID DeiT and the baseline DeiT model under attacks with escalating perturbation levels. This result demonstrates the significant advantages PIDformer has over the baseline model in terms of robustness, further confirming the benefits of our model.
\begin{table}[t!]
\vspace{-2mm}
\centering
\caption{\small 
Evaluation of PID DeiT versus Softmax DeiT on the clean ImageNet validation set, as well as under various adversarial attacks and out-of-distribution datasets.
}
    \vspace{-1em}
    \color{black}
    \begin{center}
    \scalebox{0.9}{
    \begin{tabular}{ll|cc}
    \toprule
         Attack& Metric/Model & \it Softmax DeiT  & PID DeiT (\%)  \\
        \midrule
        Clean& Top-1 Acc (\%)  &  72.17 & \bf 73.13 \\
        & Top-5 Acc (\%)  &  91.02 & \bf 91.76 \\
         \hline\midrule
        FGSM& Top-1 Acc (\%)  & 33.64  & \bf 38.52 \\
        & Top-5 Acc (\%)  & 68.18  & \bf 72.53 \\
         \hline
        PGD& Top-1 Acc (\%)  & 12.02  & \bf 15.08 \\
        & Top-5 Acc (\%)  & 34.99  & \bf 39.69 \\
         \hline
        SPSA& Top-1 Acc (\%)  & 65.75  & \bf67.98\bf  \\
        & Top-5 Acc (\%)  & 90.07  & \bf 90.58 \\
         \hline
        SLD& Top-1 Acc (\%)  & 69.32  & \bf 70.84 \\
        & Top-5 Acc (\%)  & 90.8  & \bf 91.43 \\
         \hline
        Noise & Top-1 Acc (\%)  & 69.2  & \bf 70.87 \\
        & Top-5 Acc (\%)  & 89.67  &\bf 90.77 \\
         \hline\midrule
         Imagenet-A & Top-1 Acc (\%)  & 6.90  & \bf 8.82 \\
         Imagenet-R & Top-1 Acc (\%)  & 32.83  & \bf 34.89 \\
         Imagenet-C & mCE ($\downarrow$)  & 71.20  & \bf 68.41 \\
         Imagenet-O & AUPR   & 17.47  & \bf 19.22 \\
        \bottomrule
    \end{tabular}}
    \end{center}
\vspace{-2mm}
\label{tab:attacks}
\end{table}

\textbf{Image Segmentation on ADE20K dataset.}
We evaluate the performance of Segmenter models \cite{strudel2021segmenter} using both PID DeiT and DeiT backbones on the ADE20K image segmentation task \cite{zhou2017scene}, as outlined in Table \ref{tab:segment}. The outcomes illustrate significant performance enhancements obtained by employing the PID DeiT backbone instead of the DeiT backbone across both single-scale (SS) and multi-scale (MS) Mean Intersection over Union (MIoU) metrics. 
\begin{table}[t!]
\vspace{-5mm}
\centering
\caption{\small 
\small 
Single-scale (SS) MIoU and multi-scale MIoU (MS) of the PID DeiT vs. the DeiT on the ADE20K image segmentation. 
}
    \color{black}
    \begin{center}
    \scalebox{0.9}{
    \begin{tabular}{l|cc}
    \toprule
         Model/Metric & SS MIoU  & MS MIoU (\%)  \\
        \midrule
        \it Softmax DeiT  &  35.72 & 36.68 \\
        PID DeiT & \bf 37.42 & \bf 38.28 \\
        \bottomrule
    \end{tabular}}
    \end{center}
\label{tab:segment}
\end{table}

\textbf{Language Model on WikiText-103.}
In addition to computer vision tasks, we evaluate our model's performance in the language modeling task on the WikiText-103 dataset (Table~\ref{tab:lm}). Our PIDformer language model surpasses the softmax transformer model \cite{xiong2021nystromformer} in test and valid perplexity. These results, combined with findings across various tasks, empirically demonstrate the significant advantages of PIDformer models.
\begin{table}[t!]
\vspace{-4mm}
\footnotesize
    \caption{\small Test and valid perplexity (Test PPL and Valid PPL) on WikiText-103 of PIDformer compared to the softmax transformer.
    }
    \begin{center}
    \scalebox{1.}{
    \begin{tabular}{l|cc}
    \toprule
        Method/Metric & Valid PPL & Test PPL \\
        \midrule
        {\it Softmax Transformer} & 33.15  & 34.29 \\
        PIDformer & \bf 32.44  & \bf 33.45  \\
        \bottomrule
    \end{tabular}}
    \end{center}
\label{tab:lm}
\end{table}

\begin{figure}[!t]
\vspace{-5mm}
\centering
\includegraphics[width=0.38\textwidth]{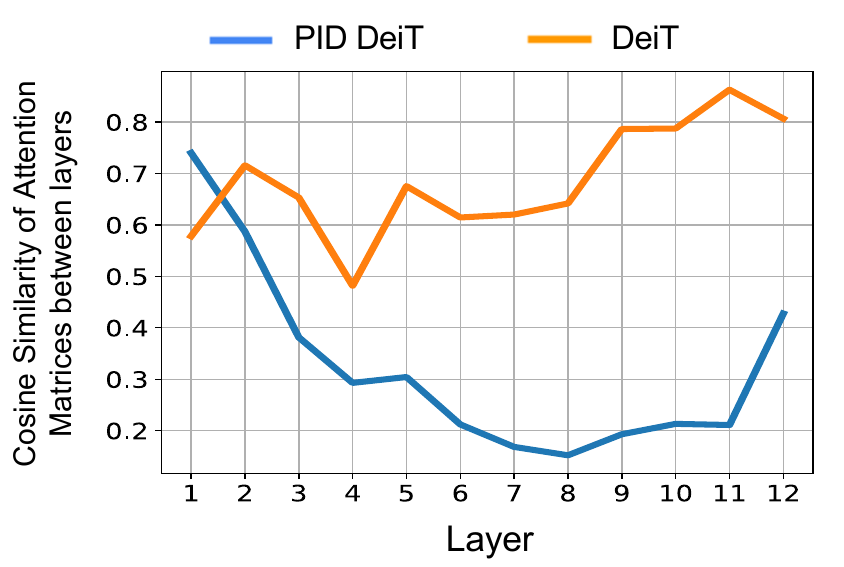}
\vspace{-3mm}
\caption{\small The cosine \textit{similarity} of token representations in PID DeiT compared to baseline DeiT models across layers for ImageNet classification. The DeiT baseline demonstrates representation rank collapse as tokens become increasingly similar as depth increases. In contrast, PID DeiT models exhibit significantly greater diversity in tokens, indicating a mitigation in rank-collapse.}
\label{fig:pid-cossim} 
\end{figure}

\textbf{Representation Collapse Analysis.}
We empirically show PIDformer's effectiveness in addressing rank collapse in transformers. In Fig.~\ref{fig:pid-cossim}, we compare token representation cosine similarity across layers in PID DeiT and softmax baseline models pretrained on Imagenet. PID DeiT exhibits significantly lower similarity, especially in later layers, alleviating rank collapse and enhancing token embedding diversity. Further details are in Appendix~\ref{secapp:cossim}.
\vspace{-3mm}
\section{Related Work}
\label{sec:related_work}
\textbf{Robust transformer.}
Ensuring the generalization and robustness of both vision transformer and language model remains an ongoing research focus. Large language models are vulnerable to input corruption~\cite{005357537ef34c158137a70c5df3799b, peyrard-etal-2022-invariant, jin2020bert, zang2019word}, posing a challenge in developing robust real-world applications that can withstand unforeseen adversarial threats. For ViTs, investigations into model robustness against adversarial attacks, domain shifts, and out-of-distribution data are crucial for real-world deployment. Techniques such as data augmentation, regularization, and adversarial training are actively explored to enhance the robustness and generalization capabilities of ViTs. Many investigations (e.g.,~\cite{yuan2023you, paul2022vision, mahmood2021robustness,
bhojanapalli2021understanding, madry2017towards, mao2022towards, zhou2022understanding}) have attempted to explain and improve the resilience of ViT models against typical adversarial attacks. For example,~\cite{mahmood2021robustness} empirically mitigates ViT's vulnerability to white-box adversarial attacks by introducing a simple ensemble defense strategy that notably enhanced robustness without sacrificing accuracy on clean data.

\textbf{Rank-collapse in transformer.} Rank collapse in deep transformers, observed across domains from natural language processing~\cite{shi2022revisiting} to computer vision~\cite{wang2022antioversmoothing,dong2021attention}, is evident. In computer vision, \citeauthor{zhou2021deepvit} (\citeyear{zhou2021deepvit}) find that adding more layers to the Vision Transformer (ViT)~\cite{DosovitskiyB0WZ21} quickly saturates its performance. Moreover, their experiments show that a 32-layer ViT performs worse than a 24-layer ViT, attributed to token representations becoming identical with increasing model depth. To address this matter, \cite{wang2022antioversmoothing} discovers that self-attention functions as a low-pass filter, causing token representations in ViTs to be smoothed.
Furthermore, \cite{shi2022revisiting} identifies a similar phenomenon in BERT~\cite{devlin2018bert}, and investigates rank-collapse from a graph perspective. The study employs hierarchical fusion techniques to retain the output of self-attention across all layers.
Our work is orthogonal to the existing method as we develop a control framework to tackle the non-robustness and rank-collapse issues in transformers. 

\section{Concluding Remarks}
\label{sec:conclusion}
In this paper, we present a novel control framework for self-attention mechanisms, revealing their inherent non-robustness and susceptibility to rank collapse in token representation. Leveraging this control perspective, we introduce the PIDformer, a novel PID-control Transformer designed to enhance robustness and mitigate the rank-collapse issue. Empirical validation across a range of large-scale applications, including ImageNet object classification (under various input perturbations and robustness benchmarks), ADE20K object segmentation, and WikiText-103 language modeling, confirms PIDformer's benefits. A limitation of our paper is the oversight regarding the privacy-preserving aspects of PIDformer. Exploring the potential of controlled transformers in enhancing privacy-preserving techniques is an intriguing avenue for future research.

\newpage
\section*{Impact Statement}
This paper presents work whose goal is to advance the field
of Machine Learning. There are many potential societal
consequences of our work, none which we feel must be
specifically highlighted here.
\bibliography{ref}

\begin{thebibliography}{56}
\providecommand{\natexlab}[1]{#1}
\providecommand{\url}[1]{\texttt{#1}}
\expandafter\ifx\csname urlstyle\endcsname\relax
  \providecommand{\doi}[1]{doi: #1}\else
  \providecommand{\doi}{doi: \begingroup \urlstyle{rm}\Url}\fi

\bibitem[Al-Rfou et~al.(2019)Al-Rfou, Choe, Constant, Guo, and Jones]{al2019character}
Al-Rfou, R., Choe, D., Constant, N., Guo, M., and Jones, L.
\newblock Character-level language modeling with deeper self-attention.
\newblock In \emph{Proceedings of the AAAI Conference on Artificial Intelligence}, volume~33, pp.\  3159--3166, 2019.

\bibitem[Bandeira et~al.(2020)Bandeira, Singer, and Strohmer]{strohmer2020mathdl}
Bandeira, A.~S., Singer, A., and Strohmer, T.
\newblock \emph{Mathematics of Data Science}.
\newblock 2020.
\newblock URL \url{https://people.math.ethz.ch/~abandeira/BandeiraSingerStrohmer-MDS-draft.pdf}.

\bibitem[Bhojanapalli et~al.(2021)Bhojanapalli, Chakrabarti, Glasner, Li, Unterthiner, and Veit]{bhojanapalli2021understanding}
Bhojanapalli, S., Chakrabarti, A., Glasner, D., Li, D., Unterthiner, T., and Veit, A.
\newblock Understanding robustness of transformers for image classification.
\newblock In \emph{Proceedings of the IEEE/CVF international conference on computer vision}, pp.\  10231--10241, 2021.

\bibitem[Chen et~al.(2021)Chen, Lu, Rajeswaran, Lee, Grover, Laskin, Abbeel, Srinivas, and Mordatch]{chen2021decision}
Chen, L., Lu, K., Rajeswaran, A., Lee, K., Grover, A., Laskin, M., Abbeel, P., Srinivas, A., and Mordatch, I.
\newblock Decision transformer: Reinforcement learning via sequence modeling.
\newblock \emph{Advances in neural information processing systems}, 34:\penalty0 15084--15097, 2021.

\bibitem[Child et~al.(2019)Child, Gray, Radford, and Sutskever]{child2019generating}
Child, R., Gray, S., Radford, A., and Sutskever, I.
\newblock Generating long sequences with sparse transformers.
\newblock \emph{arXiv preprint arXiv:1904.10509}, 2019.

\bibitem[Cho et~al.(2014)Cho, van Merri{\"e}nboer, Gulcehre, Bahdanau, Bougares, Schwenk, and Bengio]{cho-etal-2014-learning}
Cho, K., van Merri{\"e}nboer, B., Gulcehre, C., Bahdanau, D., Bougares, F., Schwenk, H., and Bengio, Y.
\newblock Learning phrase representations using {RNN} encoder{--}decoder for statistical machine translation.
\newblock In \emph{Proceedings of the 2014 Conference on Empirical Methods in Natural Language Processing ({EMNLP})}, pp.\  1724--1734, Doha, Qatar, October 2014. Association for Computational Linguistics.
\newblock \doi{10.3115/v1/D14-1179}.
\newblock URL \url{https://www.aclweb.org/anthology/D14-1179}.

\bibitem[Deng et~al.(2009)Deng, Dong, Socher, Li, Li, and Fei-Fei]{deng2009imagenet}
Deng, J., Dong, W., Socher, R., Li, L.-J., Li, K., and Fei-Fei, L.
\newblock Imagenet: A large-scale hierarchical image database.
\newblock In \emph{2009 IEEE conference on computer vision and pattern recognition}, pp.\  248--255. Ieee, 2009.

\bibitem[Devlin et~al.(2018)Devlin, Chang, Lee, and Toutanova]{devlin2018bert}
Devlin, J., Chang, M.-W., Lee, K., and Toutanova, K.
\newblock Bert: Pre-training of deep bidirectional transformers for language understanding.
\newblock \emph{arXiv preprint arXiv:1810.04805}, 2018.

\bibitem[Dong et~al.(2020)Dong, Fu, Yang, Pang, Su, Xiao, and Zhu]{dong2020benchmarking}
Dong, Y., Fu, Q.-A., Yang, X., Pang, T., Su, H., Xiao, Z., and Zhu, J.
\newblock Benchmarking adversarial robustness on image classification.
\newblock In \emph{proceedings of the IEEE/CVF conference on computer vision and pattern recognition}, pp.\  321--331, 2020.

\bibitem[Dong et~al.(2021)Dong, Cordonnier, and Loukas]{dong2021attention}
Dong, Y., Cordonnier, J.-B., and Loukas, A.
\newblock Attention is not all you need: Pure attention loses rank doubly exponentially with depth.
\newblock In \emph{International Conference on Machine Learning}, pp.\  2793--2803. PMLR, 2021.

\bibitem[Dosovitskiy et~al.(2021{\natexlab{a}})Dosovitskiy, Beyer, Kolesnikov, Weissenborn, Zhai, Unterthiner, Dehghani, Minderer, Heigold, Gelly, Uszkoreit, and Houlsby]{DosovitskiyB0WZ21}
Dosovitskiy, A., Beyer, L., Kolesnikov, A., Weissenborn, D., Zhai, X., Unterthiner, T., Dehghani, M., Minderer, M., Heigold, G., Gelly, S., Uszkoreit, J., and Houlsby, N.
\newblock An image is worth 16x16 words: Transformers for image recognition at scale.
\newblock In \emph{9th International Conference on Learning Representations, {ICLR} 2021, Virtual Event, Austria, May 3-7, 2021}. OpenReview.net, 2021{\natexlab{a}}.
\newblock URL \url{https://openreview.net/forum?id=YicbFdNTTy}.

\bibitem[Dosovitskiy et~al.(2021{\natexlab{b}})Dosovitskiy, Beyer, Kolesnikov, Weissenborn, Zhai, Unterthiner, Dehghani, Minderer, Heigold, Gelly, Uszkoreit, and Houlsby]{dosovitskiy2021an}
Dosovitskiy, A., Beyer, L., Kolesnikov, A., Weissenborn, D., Zhai, X., Unterthiner, T., Dehghani, M., Minderer, M., Heigold, G., Gelly, S., Uszkoreit, J., and Houlsby, N.
\newblock An image is worth 16x16 words: Transformers for image recognition at scale.
\newblock In \emph{International Conference on Learning Representations}, 2021{\natexlab{b}}.
\newblock URL \url{https://openreview.net/forum?id=YicbFdNTTy}.

\bibitem[Gilboa \& Osher(2007)Gilboa and Osher]{Gilboa2007NonlocalLI}
Gilboa, G. and Osher, S.
\newblock Nonlocal linear image regularization and supervised segmentation.
\newblock \emph{Multiscale Model. Simul.}, 6:\penalty0 595--630, 2007.

\bibitem[Gilboa \& Osher(2008)Gilboa and Osher]{Gilboa2008NonlocalOW}
Gilboa, G. and Osher, S.
\newblock Nonlocal operators with applications to image processing.
\newblock \emph{Multiscale Model. Simul.}, 7:\penalty0 1005--1028, 2008.

\bibitem[Gulati et~al.(2020)Gulati, Qin, Chiu, Parmar, Zhang, Yu, Han, Wang, Zhang, Wu, et~al.]{gulati2020conformer}
Gulati, A., Qin, J., Chiu, C.-C., Parmar, N., Zhang, Y., Yu, J., Han, W., Wang, S., Zhang, Z., Wu, Y., et~al.
\newblock Conformer: Convolution-augmented transformer for speech recognition.
\newblock \emph{arXiv preprint arXiv:2005.08100}, 2020.

\bibitem[Guo et~al.(2021)Guo, Cai, Liu, Mu, Martin, and Hu]{guo2021pct}
Guo, M.-H., Cai, J.-X., Liu, Z.-N., Mu, T.-J., Martin, R.~R., and Hu, S.-M.
\newblock Pct: Point cloud transformer.
\newblock \emph{Computational Visual Media}, 7\penalty0 (2):\penalty0 187--199, 2021.

\bibitem[Hendrycks \& Dietterich(2019)Hendrycks and Dietterich]{hendrycks2019benchmarking}
Hendrycks, D. and Dietterich, T.
\newblock Benchmarking neural network robustness to common corruptions and perturbations.
\newblock \emph{arXiv preprint arXiv:1903.12261}, 2019.

\bibitem[Hendrycks et~al.(2021{\natexlab{a}})Hendrycks, Basart, Mu, Kadavath, Wang, Dorundo, Desai, Zhu, Parajuli, Guo, et~al.]{hendrycks2021many}
Hendrycks, D., Basart, S., Mu, N., Kadavath, S., Wang, F., Dorundo, E., Desai, R., Zhu, T., Parajuli, S., Guo, M., et~al.
\newblock The many faces of robustness: A critical analysis of out-of-distribution generalization.
\newblock In \emph{Proceedings of the IEEE/CVF International Conference on Computer Vision}, pp.\  8340--8349, 2021{\natexlab{a}}.

\bibitem[Hendrycks et~al.(2021{\natexlab{b}})Hendrycks, Zhao, Basart, Steinhardt, and Song]{hendrycks2021natural}
Hendrycks, D., Zhao, K., Basart, S., Steinhardt, J., and Song, D.
\newblock Natural adversarial examples.
\newblock In \emph{Proceedings of the IEEE/CVF Conference on Computer Vision and Pattern Recognition}, pp.\  15262--15271, 2021{\natexlab{b}}.

\bibitem[Janner et~al.(2021)Janner, Li, and Levine]{janner2021offline}
Janner, M., Li, Q., and Levine, S.
\newblock Offline reinforcement learning as one big sequence modeling problem.
\newblock \emph{Advances in neural information processing systems}, 34:\penalty0 1273--1286, 2021.

\bibitem[Jin et~al.(2020)Jin, Jin, Zhou, and Szolovits]{jin2020bert}
Jin, D., Jin, Z., Zhou, J.~T., and Szolovits, P.
\newblock Is bert really robust? a strong baseline for natural language attack on text classification and entailment.
\newblock In \emph{Proceedings of the AAAI conference on artificial intelligence}, volume~34, pp.\  8018--8025, 2020.

\bibitem[Madry et~al.(2017)Madry, Makelov, Schmidt, Tsipras, and Vladu]{madry2017towards}
Madry, A., Makelov, A., Schmidt, L., Tsipras, D., and Vladu, A.
\newblock Towards deep learning models resistant to adversarial attacks.
\newblock \emph{arXiv preprint arXiv:1706.06083}, 2017.

\bibitem[Mahmood et~al.(2021)Mahmood, Mahmood, and Van~Dijk]{mahmood2021robustness}
Mahmood, K., Mahmood, R., and Van~Dijk, M.
\newblock On the robustness of vision transformers to adversarial examples.
\newblock In \emph{Proceedings of the IEEE/CVF International Conference on Computer Vision}, pp.\  7838--7847, 2021.

\bibitem[Mao et~al.(2022)Mao, Qi, Chen, Li, Duan, Ye, He, and Xue]{mao2022towards}
Mao, X., Qi, G., Chen, Y., Li, X., Duan, R., Ye, S., He, Y., and Xue, H.
\newblock Towards robust vision transformer.
\newblock In \emph{Proceedings of the IEEE/CVF conference on Computer Vision and Pattern Recognition}, pp.\  12042--12051, 2022.

\bibitem[Merity et~al.(2017)Merity, Xiong, Bradbury, and Socher]{DBLP:conf/iclr/MerityX0S17}
Merity, S., Xiong, C., Bradbury, J., and Socher, R.
\newblock Pointer sentinel mixture models.
\newblock In \emph{5th International Conference on Learning Representations, {ICLR} 2017, Toulon, France, April 24-26, 2017, Conference Track Proceedings}. OpenReview.net, 2017.
\newblock URL \url{https://openreview.net/forum?id=Byj72udxe}.

\bibitem[Metropolis \& Ulam(1949)Metropolis and Ulam]{13ab5b5e-0237-33fb-a7a8-6f6e4e0d4e0f}
Metropolis, N. and Ulam, S.
\newblock The monte carlo method.
\newblock \emph{Journal of the American Statistical Association}, 44\penalty0 (247):\penalty0 335--341, 1949.
\newblock ISSN 01621459.
\newblock URL \url{http://www.jstor.org/stable/2280232}.

\bibitem[Morača(2007)]{MORACA2007666}
Morača, N.
\newblock Upper bounds for the infinity norm of the inverse of sdd and s-sdd matrices.
\newblock \emph{Journal of Computational and Applied Mathematics}, 206\penalty0 (2):\penalty0 666--678, 2007.
\newblock ISSN 0377-0427.
\newblock \doi{https://doi.org/10.1016/j.cam.2006.08.013}.
\newblock URL \url{https://www.sciencedirect.com/science/article/pii/S0377042706005139}.

\bibitem[Parikh et~al.(2016)Parikh, T{\"a}ckstr{\"o}m, Das, and Uszkoreit]{parikh-etal-2016-decomposable}
Parikh, A., T{\"a}ckstr{\"o}m, O., Das, D., and Uszkoreit, J.
\newblock A decomposable attention model for natural language inference.
\newblock In \emph{Proceedings of the 2016 Conference on Empirical Methods in Natural Language Processing}, pp.\  2249--2255, Austin, Texas, November 2016. Association for Computational Linguistics.
\newblock \doi{10.18653/v1/D16-1244}.
\newblock URL \url{https://www.aclweb.org/anthology/D16-1244}.

\bibitem[Paul \& Chen(2022)Paul and Chen]{paul2022vision}
Paul, S. and Chen, P.-Y.
\newblock Vision transformers are robust learners.
\newblock In \emph{Proceedings of the AAAI conference on Artificial Intelligence}, volume~36, pp.\  2071--2081, 2022.

\bibitem[Peyrard et~al.(2022)Peyrard, Ghotra, Josifoski, Agarwal, Patra, Carignan, Kiciman, Tiwary, and West]{peyrard-etal-2022-invariant}
Peyrard, M., Ghotra, S., Josifoski, M., Agarwal, V., Patra, B., Carignan, D., Kiciman, E., Tiwary, S., and West, R.
\newblock Invariant language modeling.
\newblock In Goldberg, Y., Kozareva, Z., and Zhang, Y. (eds.), \emph{Proceedings of the 2022 Conference on Empirical Methods in Natural Language Processing}, pp.\  5728--5743, Abu Dhabi, United Arab Emirates, December 2022. Association for Computational Linguistics.
\newblock \doi{10.18653/v1/2022.emnlp-main.387}.
\newblock URL \url{https://aclanthology.org/2022.emnlp-main.387}.

\bibitem[Raffel et~al.(2020)Raffel, Shazeer, Roberts, Lee, Narang, Matena, Zhou, Li, and Liu]{JMLR:v21:20-074}
Raffel, C., Shazeer, N., Roberts, A., Lee, K., Narang, S., Matena, M., Zhou, Y., Li, W., and Liu, P.~J.
\newblock Exploring the limits of transfer learning with a unified text-to-text transformer.
\newblock \emph{Journal of Machine Learning Research}, 21\penalty0 (140):\penalty0 1--67, 2020.
\newblock URL \url{http://jmlr.org/papers/v21/20-074.html}.

\bibitem[Russakovsky et~al.(2015)Russakovsky, Deng, Su, Krause, Satheesh, Ma, Huang, Karpathy, Khosla, Bernstein, et~al.]{russakovsky2015imagenet}
Russakovsky, O., Deng, J., Su, H., Krause, J., Satheesh, S., Ma, S., Huang, Z., Karpathy, A., Khosla, A., Bernstein, M., et~al.
\newblock Imagenet large scale visual recognition challenge.
\newblock \emph{International Journal of Computer Vision}, 115\penalty0 (3):\penalty0 211--252, 2015.

\bibitem[Schlag et~al.(2021)Schlag, Irie, and Schmidhuber]{schlag2021linear}
Schlag, I., Irie, K., and Schmidhuber, J.
\newblock Linear transformers are secretly fast weight programmers.
\newblock In \emph{International Conference on Machine Learning}, pp.\  9355--9366. PMLR, 2021.

\bibitem[Shi et~al.(2022)Shi, GAO, Xu, Liang, Li, Kong, Lee, and Kwok]{shi2022revisiting}
Shi, H., GAO, J., Xu, H., Liang, X., Li, Z., Kong, L., Lee, S. M.~S., and Kwok, J.
\newblock Revisiting over-smoothing in {BERT} from the perspective of graph.
\newblock In \emph{International Conference on Learning Representations}, 2022.
\newblock URL \url{https://openreview.net/forum?id=dUV91uaXm3}.

\bibitem[Silvester(2000)]{0826aabe-2dc2-393a-bbf2-a98a36b6bed5}
Silvester, J.~R.
\newblock Determinants of block matrices.
\newblock \emph{The Mathematical Gazette}, 84\penalty0 (501):\penalty0 460--467, 2000.
\newblock ISSN 00255572.
\newblock URL \url{http://www.jstor.org/stable/3620776}.

\bibitem[Strang(2006)]{strang2006linear}
Strang, G.
\newblock \emph{Linear algebra and its applications}.
\newblock Thomson, Brooks/Cole, Belmont, CA, 2006.
\newblock ISBN 0030105676 9780030105678 0534422004 9780534422004.
\newblock URL \url{http://www.amazon.com/Linear-Algebra-Its-Applications-Edition/dp/0030105676}.

\bibitem[Strudel et~al.(2021)Strudel, Garcia, Laptev, and Schmid]{strudel2021segmenter}
Strudel, R., Garcia, R., Laptev, I., and Schmid, C.
\newblock Segmenter: Transformer for semantic segmentation.
\newblock In \emph{Proceedings of the IEEE/CVF international conference on computer vision}, pp.\  7262--7272, 2021.

\bibitem[Touvron et~al.(2021)Touvron, Cord, Douze, Massa, Sablayrolles, and Jegou]{touvron2020deit}
Touvron, H., Cord, M., Douze, M., Massa, F., Sablayrolles, A., and Jegou, H.
\newblock Training data-efficient image transformers distillation through attention.
\newblock In Meila, M. and Zhang, T. (eds.), \emph{Proceedings of the 38th International Conference on Machine Learning}, volume 139 of \emph{Proceedings of Machine Learning Research}, pp.\  10347--10357. PMLR, 18--24 Jul 2021.
\newblock URL \url{https://proceedings.mlr.press/v139/touvron21a.html}.

\bibitem[Tramer \& Boneh(2019{\natexlab{a}})Tramer and Boneh]{NEURIPS2019_5d4ae76f}
Tramer, F. and Boneh, D.
\newblock Adversarial training and robustness for multiple perturbations.
\newblock In Wallach, H., Larochelle, H., Beygelzimer, A., d\textquotesingle Alch\'{e}-Buc, F., Fox, E., and Garnett, R. (eds.), \emph{Advances in Neural Information Processing Systems}, volume~32. Curran Associates, Inc., 2019{\natexlab{a}}.
\newblock URL \url{https://proceedings.neurips.cc/paper_files/paper/2019/file/5d4ae76f053f8f2516ad12961ef7fe97-Paper.pdf}.

\bibitem[Tramer \& Boneh(2019{\natexlab{b}})Tramer and Boneh]{tramer2019adversarial}
Tramer, F. and Boneh, D.
\newblock Adversarial training and robustness for multiple perturbations.
\newblock \emph{Advances in neural information processing systems}, 32, 2019{\natexlab{b}}.

\bibitem[Uesato et~al.(2018)Uesato, O'Donoghue, Kohli, and van~den Oord]{pmlr-v80-uesato18a}
Uesato, J., O'Donoghue, B., Kohli, P., and van~den Oord, A.
\newblock Adversarial risk and the dangers of evaluating against weak attacks.
\newblock In Dy, J. and Krause, A. (eds.), \emph{Proceedings of the 35th International Conference on Machine Learning}, volume~80 of \emph{Proceedings of Machine Learning Research}, pp.\  5025--5034. PMLR, 10--15 Jul 2018.
\newblock URL \url{https://proceedings.mlr.press/v80/uesato18a.html}.

\bibitem[Vaswani et~al.(2017)Vaswani, Shazeer, Parmar, Uszkoreit, Jones, Gomez, Kaiser, and Polosukhin]{vaswani2017attention}
Vaswani, A., Shazeer, N., Parmar, N., Uszkoreit, J., Jones, L., Gomez, A.~N., Kaiser, {\L}., and Polosukhin, I.
\newblock Attention is all you need.
\newblock In \emph{Advances in neural information processing systems}, pp.\  5998--6008, 2017.

\bibitem[Wang et~al.(2021)Wang, Wang, Cheng, Gan, Jia, Li, and Liu]{005357537ef34c158137a70c5df3799b}
Wang, B., Wang, S., Cheng, Y., Gan, Z., Jia, R., Li, B., and Liu, J.
\newblock Infobert: Improving robustness of language models from an information theoretic perspective.
\newblock 2021.
\newblock Publisher Copyright: {\textcopyright} 2021 ICLR 2021 - 9th International Conference on Learning Representations. All rights reserved.; 9th International Conference on Learning Representations, ICLR 2021 ; Conference date: 03-05-2021 Through 07-05-2021.

\bibitem[Wang et~al.(2022)Wang, Zheng, Chen, and Wang]{wang2022antioversmoothing}
Wang, P., Zheng, W., Chen, T., and Wang, Z.
\newblock Anti-oversmoothing in deep vision transformers via the fourier domain analysis: From theory to practice.
\newblock In \emph{International Conference on Learning Representations}, 2022.
\newblock URL \url{https://openreview.net/forum?id=O476oWmiNNp}.

\bibitem[Wang \& Bansal(2018)Wang and Bansal]{wang-bansal-2018-robust}
Wang, Y. and Bansal, M.
\newblock Robust machine comprehension models via adversarial training.
\newblock In Walker, M., Ji, H., and Stent, A. (eds.), \emph{Proceedings of the 2018 Conference of the North {A}merican Chapter of the Association for Computational Linguistics: Human Language Technologies, Volume 2 (Short Papers)}, pp.\  575--581, New Orleans, Louisiana, June 2018. Association for Computational Linguistics.
\newblock \doi{10.18653/v1/N18-2091}.
\newblock URL \url{https://aclanthology.org/N18-2091}.

\bibitem[Xiong et~al.(2021)Xiong, Zeng, Chakraborty, Tan, Fung, Li, and Singh]{xiong2021nystromformer}
Xiong, Y., Zeng, Z., Chakraborty, R., Tan, M., Fung, G., Li, Y., and Singh, V.
\newblock {Nystr{\"o}mformer: A Nystr{\"o}m-based Algorithm for Approximating Self-Attention}.
\newblock 2021.

\bibitem[Yin et~al.(2008)Yin, Osher, Goldfarb, and Darbon]{stanbregman}
Yin, W., Osher, S., Goldfarb, D., and Darbon, J.
\newblock Bregman iterative algorithms for l(1)-minimization with applications to compressed sensing.
\newblock \emph{Siam Journal on Imaging Sciences - SIAM J IMAGING SCI}, 1, 01 2008.
\newblock \doi{10.1137/070703983}.

\bibitem[Yuan et~al.(2023)Yuan, Zhou, Zou, and Cheng]{yuan2023you}
Yuan, Z., Zhou, P., Zou, K., and Cheng, Y.
\newblock You are catching my attention: Are vision transformers bad learners under backdoor attacks?
\newblock In \emph{Proceedings of the IEEE/CVF Conference on Computer Vision and Pattern Recognition}, pp.\  24605--24615, 2023.

\bibitem[Zang et~al.(2019)Zang, Qi, Yang, Liu, Zhang, Liu, and Sun]{zang2019word}
Zang, Y., Qi, F., Yang, C., Liu, Z., Zhang, M., Liu, Q., and Sun, M.
\newblock Word-level textual adversarial attacking as combinatorial optimization.
\newblock \emph{arXiv preprint arXiv:1910.12196}, 2019.

\bibitem[Zhang et~al.(2019)Zhang, Yao, Sun, and Tay]{zhang2019deep}
Zhang, S., Yao, L., Sun, A., and Tay, Y.
\newblock Deep learning based recommender system: A survey and new perspectives.
\newblock \emph{ACM Computing Surveys (CSUR)}, 52\penalty0 (1):\penalty0 1--38, 2019.

\bibitem[Zhang et~al.(2010)Zhang, Burger, Bresson, and Osher]{zhangnonlocalbregman}
Zhang, X., Burger, M., Bresson, X., and Osher, S.
\newblock Bregmanized nonlocal regularization for deconvolution and sparse reconstruction.
\newblock \emph{SIAM Journal on Imaging Sciences}, 3\penalty0 (3):\penalty0 253--276, 2010.
\newblock \doi{10.1137/090746379}.
\newblock URL \url{https://doi.org/10.1137/090746379}.

\bibitem[Zhao et~al.(2021)Zhao, Jiang, Jia, Torr, and Koltun]{zhao2021point}
Zhao, H., Jiang, L., Jia, J., Torr, P.~H., and Koltun, V.
\newblock Point transformer.
\newblock In \emph{Proceedings of the IEEE/CVF International Conference on Computer Vision}, pp.\  16259--16268, 2021.

\bibitem[Zhou et~al.(2017)Zhou, Zhao, Puig, Fidler, Barriuso, and Torralba]{zhou2017scene}
Zhou, B., Zhao, H., Puig, X., Fidler, S., Barriuso, A., and Torralba, A.
\newblock Scene parsing through ade20k dataset.
\newblock In \emph{Proceedings of the IEEE conference on computer vision and pattern recognition}, pp.\  633--641, 2017.

\bibitem[Zhou et~al.(2018)Zhou, Zhao, Puig, Xiao, Fidler, Barriuso, and Torralba]{zhou2018semantic}
Zhou, B., Zhao, H., Puig, X., Xiao, T., Fidler, S., Barriuso, A., and Torralba, A.
\newblock Semantic understanding of scenes through the ade20k dataset, 2018.

\bibitem[Zhou et~al.(2021)Zhou, Kang, Jin, Yang, Lian, Jiang, Hou, and Feng]{zhou2021deepvit}
Zhou, D., Kang, B., Jin, X., Yang, L., Lian, X., Jiang, Z., Hou, Q., and Feng, J.
\newblock Deepvit: Towards deeper vision transformer, 2021.

\bibitem[Zhou et~al.(2022)Zhou, Yu, Xie, Xiao, Anandkumar, Feng, and Alvarez]{zhou2022understanding}
Zhou, D., Yu, Z., Xie, E., Xiao, C., Anandkumar, A., Feng, J., and Alvarez, J.~M.
\newblock Understanding the robustness in vision transformers.
\newblock In \emph{International Conference on Machine Learning}, pp.\  27378--27394. PMLR, 2022.

\end{thebibliography}
\bibliographystyle{icml2024}

\newpage
\appendix
\onecolumn



\label{sec:appendix}
\section{Additional Details on the Experiments in Section~\ref{sec:experiment}}
\label{secapp:experiments}
This section provides datasets, models, and training details for experiments in Section~\ref{sec:experiment}. 
\subsection{Image Classification on Imagenet}
\textbf{Datasets and Metrics.} The ImageNet dataset, as described in~\cite{deng2009imagenet, russakovsky2015imagenet}, consists of $1.28$ million images for training and $50,000$ images for validation, covering the classification of 1000 different categories. Performance evaluation is based on top-1 and top-5 accuracies. \\\\
\textbf{Models and Baselines.} Our baseline model is the DeiT-tiny model~\cite{touvron2020deit}, which consists of 12 transformer layers, 3 attention heads per layer, and a model dimension of 192. For model setting and setting and configuration, we follow~\cite{touvron2020deit}. Their implementation is available at \href{https://github.com/facebookresearch/deit}{https://github.com/facebookresearch/deit}. The $\lambda_P, \lambda_I, \lambda_D$, and $\beta$ used for our PID DeiT method is $0.8, 0.5, 0.05$, and $0.1$, respectively.
\subsection{Image Segmentation on ADK20 dataset}
\textbf{Datasets and Metrics.} 
The ADE20K dataset is renowned for its incorporation of complex scenes featuring detailed labels, establishing it as one of the most rigorous semantic segmentation datasets. It comprises a training set of $20,210$ images covering 150 semantic classes. Furthermore, the dataset includes $2,000$ images in the validation set and $3,352$ images in the test set. Performance in this task is evaluated using the Single-scale mean Intersection over Union (SS mIoU) and Multi-scale (MS mIoU) metrics.

\textbf{Models and baselines.}
The training configuration and setting for our models are followed by~\cite{strudel2021segmenter}. The baseline model is finetuned with the pretrained DeiT-tiny backbone while our segmenter model used the pretrained PID DeiT-tiny, with $\lambda_P, \lambda_I, \lambda_D$, and $\beta$ are $0.5, 0.3, 0.05$, and $1$, respectively.
\subsection{Language Modeling on WikiText-103}
\textbf{Datasets and Metrics.} 
The WikiText-103 dataset is composed of Wikipedia articles tailored to capture extensive contextual dependencies. Its training set includes roughly $28,000$ articles, totaling around 103 million words. Each article consists of text blocks averaging about $3,600$ words. The validation and test sets contain $218,000$ and $246,000$ words, respectively, divided into 60 articles per set and approximately $268,000$ words each. Our experiment adheres to the standard setup outlined in~\cite{DBLP:conf/iclr/MerityX0S17, schlag2021linear}, which entails segmenting the training data into independent long segments of length $L$ words. For evaluation, we utilize a batch size of $1$ and process the text sequence using a sliding window of size $L$. When calculating perplexity (PPL), we only consider the last position, except for the first segment where all positions are evaluated, consistent with the methodology in~\cite{al2019character, schlag2021linear}.

\textbf{Models and baselines.} For our language modeling implementation, we rely on the publicly available code \textcolor{blue}{\href{https://github.com/IDSIA/lmtool-fwp}{https://github.com/IDSIA/lmtool-fwp}} developed by~\cite{schlag2021linear}. In our experiments, we set the dimensions of keys, values, and queries to 128, while the training and evaluation context length is set to 256. In this experiment, $\lambda_P, \lambda_I, \lambda_D$, and $\beta$ being set to $0.4, 0.5, 0.1$ and $0.3$, respectively, yields the best performance of PIDformer language model.
\subsection{Adversarial Examples and  Out-of-distribution datasets}
\textbf{Imagenet-C} To assess robustness against typical image corruptions, we employ the ImageNet-C dataset~\cite{hendrycks2019benchmarking}, which comprises 15 categories of artificially generated corruptions spanning five levels of severity. ImageNet-C evaluates models using the mean corruption error (mCE) metric, where a lower mCE value indicates greater resilience to corruption.

\textbf{Imagenet-A} ImageNet-A~\cite{hendrycks2021natural} is a dataset consisting of authentic images that have been filtered to deceive ImageNet classifiers. Specifically, it focuses on a subset of 200 classes chosen from the original 1000 classes in ImageNet-1K. Errors made within these 200 classes are regarded as significant, encompassing a wide range of categories represented in ImageNet-1K.

\textbf{Imagenet-O} This dataset comprises examples that have been adversarially filtered to challenge out-of-distribution detectors for ImageNet~\cite{hendrycks2021natural}. It includes samples from the larger ImageNet-22K dataset but excludes those from ImageNet1K. Specifically, samples are chosen if they are confidently misclassified as an ImageNet-1K class by a ResNet-50 model. The evaluation metric utilized is the area under the precision-recall curve (AUPR).

\textbf{Imagenet-R} This dataset comprises diverse artistic interpretations of object classes found in the original ImageNet dataset, a practice discouraged by the creators of the original ImageNet~\cite{hendrycks2021many}. ImageNet-R encompasses 30,000 renditions of images representing 200 classes from the ImageNet dataset, with a selection made from a subset of the ImageNet-1K classes.
\subsection{Adversarial Attacks}
We employ fast gradient sign method (FGSM)~\cite{dong2020benchmarking}, projected gradient descent method (PGD)~\cite{tramer2019adversarial}; and Sparse $L_1$ descent method as well as noise-adding attack
These attacks were
applied to the entire validation set of ImageNet. FGSM and PGD attacks distort the input image with a perturbation budget $\epsilon= 3/255$, and $\epsilon=0.1$ for SPSA, under $l_{\infty}$ norm, while the PGD attack uses 20 steps with a step size of $\alpha = 0.15$. For the SLD and noise attack, we follow the same setting in~\href{https://github.com/cleverhans-lab}{https://github.com/cleverhans-lab}
\subsection{Rank-collapse Analysis}
\label{secapp:cossim}
The average cosine similarity between all pairs of token's representations $(\bx_i, \bx_j)$ in a sequence is computed as
\begin{align}
\nonumber
\frac{1}{N(N - 1)}\sum_{i \neq j}\frac{\bx_i^T\bx_j}{\| \bx_i\|_2\| \bx_j\|_2}.
\end{align} 
The result is then averaged over 1000 randomly chosen test data in ImageNet. The result is then reported for each layer, as in Fig.~\ref{fig:pid-cossim}.
\section{Technical Proofs}
\subsection{Solution of the first order ODE}
\label{secapp:first-order-ode}
Given $\bm{Q} \in \mathbb{R}^{n \times n}$, $\bm{Y}(t) \in \mathbb{R}^{N \times P}, t > 0$, we are interested in the solution of the first order ODE stated as:
\begin{align}
\label{eq:ode}
    \bm{Y}'(t) = \bm{Q}\bm{Y}(t), \bm{Y}(0) = \bm{Y}^0.
\end{align}
The general solution of~(\ref{eq:ode}) is $\bm{Y}(t) = \mathrm{exp}(\bm{Q}t)\bm{C}$, where $\bm{C} \in \mathbb{R}^{n \times p}$ is an any constant matrix. Indeed, 
\begin{equation}
    \begin{aligned}
        \bm{Y}'(t) &= (\bm{I} + \bm{Q}t + \frac{\bm{Q}^2t^2}{2!} + \frac{\bm{Q}^3t^3}{3!} + \dots)'\bm{C}\\
        &= (\bm{Q} + \bm{Q}^2t + \frac{\bm{Q}^3t}{2!} + \dots)\bm{C}\\
        & = \bm{Q}\mathrm{exp}(\bm{Q}t)\bm{C} = \bm{Q}\bm{Y}(t).
    \end{aligned}
\end{equation}\\\\
To satisfy the intitial condition, $\bm{Y}(0) = \bm{Q}\mathrm{exp}(\bm{Q}0)\bm{C} = \bm{Y}^0$. Hence, $\bm{C} = \bm{Y}^0$ and the solution for the intial value problem in~(\ref{eq:ode}) is $\mathrm{exp}(\bm{Q}t)\bm{Y}^0$.
\\  \\
Every square matrix can be Jordan decomposed as the form of $\bm{Q} = \bm{S}\bm{J}\bm{S}^{-1}$, where $\bm{S}$ is invertible and contains the generalized eigenvectors of $\bm{Q}$, and $\bm{J} = \bm{\mathrm{diag}}(\bm{J}_{\eta_1, m_1}, \bm{J}_{\eta_2, m_2}, \dots, \bm{J}_{\eta_M, m_M}) $ is the Jordan form of matrix $\bm{Q}$ with,\\
$\bm{J}_{\eta_i, m_i} = \begin{bmatrix} 
    \eta_i &  1& \dots &0\\
   \vdots & \ddots & & \vdots\\
    &  & \eta_i& 1\\
    0 & \dots    &   & \eta_i
    \end{bmatrix} \in \mathbb{R}^{m_i \times m_i}$, for $i = 1, \dots, M$ are Jordan blocks and $\eta_1, \dots \eta_M$ are eigenvalues of $\bm{Q}$.
\\
We rewrite the solution of~(\ref{eq:ode}) using the Jordan decomposition as
\begin{equation}
\label{eq:ode-sol}
    \begin{aligned}
        \bm{Y}(t) &= \mathrm{exp}(\bm{Q}t)\bm{Y}^0 = \mathrm{exp}(\bm{S}\bm{J}\bm{S}^{-1}t)\bm{Y}^0\\
        & = (\bm{S}\bm{S}^{-1} + \bm{S}\bm{J}\bm{S}^{-1}t + \frac{(\bm{S}\bm{J}\bm{S}^{-1})^2t^2}{2!} + \dots)\bm{Y}^0\\
        & = \bm{S}\mathrm{exp}(\bm{J}t)\bm{S}^{-1}\bm{Y}^0.
    \end{aligned}
\end{equation}
\\
We are now interested in the asymptotic behavior of the solution in~(\ref{eq:ode-sol}) as $t \rightarrow \infty$.\\
\textit{\textbf{When $\bm{Q}$ only has eigenvalues negative real parts}}. As $\eta_1, \dots \eta_M < 0$, we consider
\begin{equation}
    \begin{aligned}
        \mathrm{exp}(\bm{J}_{\eta_i, m_i}t) &= \sum_{k = 0}^{\infty}\frac{(\bm{J}_{\eta_i, m_i}t)^k}{k!}\\
        & = \begin{bmatrix}
            \displaystyle \sum_{k = 0}^{\infty} \frac{t^k\eta_i^{k}}{k!} & \displaystyle \sum_{k = 1}^{\infty} \frac{t^k\eta_i^{k - 1}}{(k - 1)!} & \dots & \displaystyle \sum_{k = m_i}^{\infty} \frac{t^k\eta_i^{k - m_i + 1}}{(k - m_i + 1)!}\\
            \vdots & \ddots & & &\\
            0&\dots & \displaystyle \sum_{k = 0}^{\infty} \frac{t^k\eta_i^{k}}{k!} & \displaystyle \sum_{k = 1}^{\infty} \frac{t^k\eta_i^{k - 1}}{(k - 1)!}\\
            0& \dots & 0 & \displaystyle \sum_{k = 0}^{\infty} \frac{t^k\eta_i^{k}}{k!}
        \end{bmatrix}\\
        &= \begin{bmatrix}
            e^{\eta_it} & te^{\eta_it} & \dots & t^{m_i - 1}e^{\eta_it}\\
            \vdots & \ddots & & &\\
            0&\dots & e^{\eta_it} & te^{\eta_it}\\
            0& \dots & 0 & e^{\eta_it}
        \end{bmatrix}\\
    \end{aligned}
\end{equation}
which is derived from the result $\bm{J}^k_{\eta_i, m_i} = \begin{bmatrix}
             \eta_i^{k} & \begin{pmatrix} j\\1 \end{pmatrix}\eta_i^{k - 1} & \dots & \begin{pmatrix} j\\m_i - 1 \end{pmatrix}\eta_i^{k - m_i + 1}\\
            \vdots & \ddots & & &\\
            0&\dots & \eta_i^{k} & \begin{pmatrix} j\\1 \end{pmatrix}\eta_i^{k - 1}\\
            0& \dots & 0 & \eta_i^{k}
    \end{bmatrix}\\$  
Therefore, when $t \rightarrow 0, \mathrm{exp}(\bm{J}_{\eta_i, m_i}t)\rightarrow \bm{0}$, making $\mathrm{exp}(\bm{J}t) \rightarrow \bm{0}$ and hence the solution in~(\ref{eq:ode-sol}) will goes to $\bm{0}$ or being stable.\\
\textit{\textbf{When $\bm{Q}$ only has at least one eigenvalue with positive real part}}. Without the loss of generalization, let $\mathrm{Re}(\eta_1) > 0$. Hence $\| \mathrm{exp}(\bm{J}_{\eta_1, m_i}t) \| \rightarrow \infty$ when $t \rightarrow \infty$. In other words, the solution of~(\ref{eq:ode}) will explode or unstable.
\subsection{Proof of Lemma~\ref{lem:state-space-sol}}
\label{secapp:state-space-sol}
The proof of Lemma~\ref{lem:state-space-sol} is the direct result in Appendix~\ref{secapp:first-order-ode}. The solution of the ordinary differential equation (ODE) in~(\ref{eq:ode1}) is
        $\bV(t) = \bm{P}\mathrm{exp}(\bm{J}t)\bm{P}^{-1}\bV^0$,
where $\bm{P}\bm{J}\bm{P}^{-1}$ if the Jordan decomposition of $\bm{K} - \bm{I}$, $\bm{J} = \bm{\mathrm{diag}}(\bm{J}_{\alpha_1, m_1}, \bm{J}_{\alpha_2, m_2}, \dots, \bm{J}_{\alpha_M, m_M})$ and $\alpha_1 \geq \alpha_2 \geq \dots, \geq \alpha_M, M \leq N$ are  eigenvalues $\bK - \bI$. Consequently, we have proved the Lemma~\ref{lem:state-space-sol}
\subsection{Proof of Lemma~\ref{lem:steady-state-state-space-sol}}
\label{secapp:steady-state-state-space-sol}
In Section~\ref{subsec:state-space-analysis}, we have shown that $\bm{K} - \bm{I}$ has a unique largest eigenvalue $\lambda_1 = 0$. This means that the Jordan blocks corresponding with other eigenvalues which has negative real parts will approach $\mathrm{exp}(\bm{J}_{\eta_i, m_i}t) \rightarrow \bm{0}$, for $i = 1, \dots, M;, i \neq 1$, as $t \rightarrow \infty$. As the consequence, $\mathrm{exp}(\bm{J}t)$ are fill with 0 for all entries except for the first entry $\mathrm{exp}(\bm{J}t)(0, 0) = 1$. Hence, the solution in~(\ref{eq:state-space-sol}) becomes 
\begin{align}
\nonumber
    \begin{bmatrix} \displaystyle c_{1,1}\bm{p_1}, & \dots ,& \displaystyle c_{1, D_x}\bm{p_1} \end{bmatrix}.
\end{align}
This concludes the proof.
\subsection{Proof of Lemma~\ref{lem:breg-iter}}
\label{secapp:breg-iter}
For $\bv^{\ell + 1}$ to be the solution of the optimization problem in~(\ref{eqn:breg-op}), since $\bm{0} \in \partial J(\bv^{\ell + 1}) - \bm{p}^{\ell} + \partial G(\bv^{\ell + 1}, \bff)$, hence the iteration becomes:
\begin{equation}
\nonumber
\begin{cases} 
& \bv^{\ell + 1} = \displaystyle \argmin_{\bv} J(\bv) - <\bm{p}^{\ell}, \bv> + G(\bv, \bff) \\
& \bm{p}^{\ell + 1} \in \bm{p}^{\ell} - \partial G(\bv^{\ell + 1}, \bff). 
\end{cases}
\end{equation}
When $G(\bv, \bff) = \frac{\lambda}{2}  \int_{\Omega} \|\bv(x) - \bff(x)\|_2^2 dx$,
\begin{equation}
\nonumber
\begin{aligned}
G(\bv, \bff) - \langle\bm{p}^{\ell}, \bv\rangle &= \frac{\lambda}{2} \int_{\Omega}\Bigg(  \left(  \|\bv(x)\|_2^2 -2\langle\bv(x), \bff(x)\rangle + \|\ \bff(x)\|_2^2 \right ) + \lambda\langle\sum^{\ell}_{m = 1} \left(\bv^m(x) - \bff(x) \right), \bv(x)\rangle \Bigg) dx \\
&= \frac{\lambda}{2} \int_{\Omega} \Bigg( \|\bv(x)\|_2^2 - \lambda\langle\bff(x) - \sum^{\ell}_{m = 1}\Big( \bv^m(x) - \bff(x) \Big), \bv(x)\rangle\Bigg) dx + \text{constant}\\
&= \frac{\lambda}{2} \int_{\Omega} \|\bv(x) - \bff^{\ell}(x)\|_2^2 dx + \text{constant},
\end{aligned}
\end{equation}
where $\bff^{\ell}(x) = \bff^{\ell - 1}(x) + \bff(x) - \bv^{\ell}(x)$. \\
Substituting $G(\bv, \bff) - \langle\bm{p}^{\ell}, \bv\rangle$ into the iteration, it becomes
\begin{equation}
\begin{cases}
   & \bv^{\ell + 1} = \displaystyle \argmin_{\bv} J(\bv) + \frac{\lambda}{2} \int_{\Omega} \|\bv(x) - \bff^{\ell}(x)\|_2^2 dx \\
 & \bff^{\ell}(x) = \bff^{\ell - 1}(x) + \bff(x) - \bv^{\ell}(x).
    \end{cases}
\end{equation}
The iteration in Lemma~\ref{lem:breg-iter} can be reformulated as:
\begin{equation}
\nonumber
   \bv^{\ell + 1} = \displaystyle \argmin_{\bv} J(\bv) + \frac{\lambda}{2} \int_{\Omega} \|\bv(x) - \bff(x) - \bm{e}_{\mathrm{a}}^{\ell}(x)\|_2^2 dx \\
\end{equation}
where $\bm{e}^{\ell}_{\mathrm{a}}(x)= \sum_{m = 1}^{\ell} \bm{e}^m(x) = \sum_{m = 1}^{\ell}\big( \bff(x) - \bv^m(x) \big)$
we conclude the proof for Lemma~\ref{lem:breg-iter}.
\subsection{Proof of Lemma~\ref{lem:state-space-P-sol}}
\label{secapp:state-space-P-sol}
To find the solution of Eqn~\ref{eq:odeP}, firstly, we find the solution for the homogenous ODE:
\begin{align}
    \bV^{(h)}{'}(t) = \bigl(\bK - (\lambda_P + 1)\bm{I} \bigl)\bV^{(h)}(t)
\end{align}
From the result in Appendix~\ref{secapp:first-order-ode}, the solution for this ODE is $\mathrm{exp}(\bm{B}t)\bm{C}$ where $\bm{B} = \bK - (\lambda_P + 1)\bm{I}$ and $\bm{C} \in \mathbb{R}^{N \times D_x} $ is any constant matrix.
Secondly, we find a particular solution for~(\ref{eq:odeP}) by solving $\bV^{(p)}{'}(t) = \bm{B}\bV(t)^{(p)} + \lambda_P\bm{F} = \bm{0}$. Since $\bm{B}$ is invertible, the solution for this equation is $\bV^{(p)}(t) = -\lambda_P\bm{B}^{-1}\bm{F}$. It is easy to check that $\bV(t) = \bV^{(h)}(t) + \bV^{(p)}(t)$ is the solution of the $\bV'(t) = \bm{B}\bV(t) + \lambda_P\bm{F}$. Applying the initial condition, $\bV(0) = \bm{C} -\lambda_P\bm{B}^{-1}\bm{F} = \bV^0$, we find $\bm{C} = \bV^0 + \lambda_P\bm{B}^{-1}\bm{F}$. Therefore, we have proved that the solution for the IVP problem in~(\ref{eq:odeP}) is indeed $\bV(t) = \mathrm{exp}(\bm{B}t)(\bV^0 + \bm{B}^{-1}\bm{F}) - \lambda_P\bm{B}^{-1}\bm{F}.$\\ \\
In Section~\ref{subsec:PD-robust}, we show that $\bm{B}$ has only eigenvalues with negative real parts. As the result in Appendix~\ref{secapp:first-order-ode}, when $t \rightarrow 0$, the $\mathrm{exp}(\bm{B}t) \rightarrow \bm{0}$ , leading to the vanishing of the $\bV^{(h)}(t)$. Hence the steady state solution for the ODE in~(\ref{eq:odeP}) becomes $ - \lambda_P\bm{B}^{-1}\bm{F}$.
\\ \\
This concludes the proof.
\subsection{Proof of Proposition~\ref{lem:lambda_P}}
\label{secapp:lambda_P}
We first show that $\bm{B}$ is a strictly diagonal dominant (SDD) matrix, i.e., $|\bm{B}(i, i)| > |\sum_{j \neq i}^N\bm{B}(i, j)|$, for $i, j = 1, \dots, N$. In fact, $|\bm{B}(i, i)| = |\bm{K}(i, i) - \lambda_p - 1| > |1 - \bm{K}(i, i)| = |\sum_{j \neq i}^N\bm{K}(i, j)| = |\sum_{j \neq i}^N\bm{B}(i, j)|$ because $\bm{K}$ is a right-stochastic matrix with all entries in $(0, 1]$ and sum of each row is 1.
\\
Hence, following~\cite{MORACA2007666}, the upper bound of $\| \bm{B}^{-1}\|_{\infty}$, when $\bm{B}$ is an SDD matrix, is given as
\begin{align}
    \| \bm{B}^{-1}\|_{\infty} &\leq \frac{1}{ \displaystyle \min_{i \in N}(|\bm{B}(i, i)| - |\sum_{j \neq i}^N\bm{B}(i, j)|)} \\
    & = \frac{1}{|\bm{K}(i, i) - \lambda_p - 1| - |1 - \bm{K}(i, i)|} = \frac{1}{\lambda_P},
\end{align} where $\displaystyle \| \bm{B}^{-1}\|_{\infty} = \max_{i = 1}^N \sum_{j = 1}^N |\bm{B}^{-1}(i, j)|$.\\
On the other hand, 
\begin{equation}
    \begin{aligned}
    \|-\lambda_P\beta \bm{B}^{-1}\bm{\epsilon}\|_{\infty} &\leq \lambda_P\beta \| \bm{B}^{-1}\|_{\infty}\| \bm{\epsilon}\|_{\infty} \\
    &= \lambda_P \beta \frac{1}{\lambda_P} \bar{\epsilon}
    =\beta\bar{\epsilon}
    \end{aligned}
\end{equation}
For the bounded error get arbitrarily small, we constraint $\beta\bar{\epsilon} \leq \delta$, making $\beta \leq \frac{\delta}{\bar{\epsilon}}$.\\
Here in the proof, we used the submultiplicity property of $\|.\|_{\infty}$ norm of matrices, which is proved as follow:
\begin{equation}
    \begin{aligned}
        \nonumber
        \| \bm{B}^{-1}\bm{\epsilon}\|_{\infty} = \sup_{\bx}\frac{\|\bm{B}^{-1}\bm{\epsilon} \bx\|_{\infty}}{\| \bx\|_{\infty}} &= \sup_{\bx}\frac{\|\bm{B}^{-1}\bm{\epsilon} \bx\|_{\infty} \|\bm{\epsilon} \bx\|_{\infty}}{\|\bm{\epsilon} \bx\|_{\infty}\| \bx\|_{\infty}}\\
        &\leq \sup_{\bx}\frac{\|\bm{B}^{-1}\bm{\epsilon} \bx\|_{\infty}}{\| \bm{\epsilon} \bx\|_{\infty}} \sup_{\bx}\frac{\|\bm{\epsilon} \bx\|_{\infty}}{\| \bx\|_{\infty}}\\
        & \leq \sup_{\bx}\frac{\|\bm{ B^{-1}} \bx\|_{\infty}}{\| \bx\|_{\infty}}\sup_{\bx}\frac{\|\bm{\epsilon} \bx\|_{\infty}}{\| \bx\|_{\infty}}\\
        &= \|\bm{B}^{-1}\|_{\infty}\|\bm{\epsilon}\|_{\infty}
    \end{aligned}
\end{equation}
With this, we conclude the proof of Proposition~\ref{lem:lambda_P}
\subsection{Proof of Lemma~\ref{lem:state-space-PD-sol}}
\label{secapp:state-space-PD-sol}
To find the solution of~(\ref{eq:odePD}), firstly, we find the solution for the homogenous ODE:
\begin{align}
\nonumber
    \bV^{(h)}{'}(t) = \frac{1}{1 + \lambda_D}\bigl(\bK - (\lambda_P + 1)\bm{I} \bigl)\bV^{(h)}(t)
\end{align}
From the result in Appendix~\ref{secapp:first-order-ode}, the solution for this ODE is $\mathrm{exp}(\displaystyle \frac{1}{\lambda_D + 1}\bm{B}t)\bm{C}$ where $\bm{B} = \bK - (\lambda_P + 1)\bm{I}$ and $\bm{C} \in \mathbb{R}^{N \times D_x} $ is any constant matrix.
Secondly, we find a particular solution for~(\ref{eq:odePD}) by solving $\bV^{(p)}{'}(t) = \displaystyle \frac{1}{\lambda_D + 1}(\bm{B}\bV(t)^{(p)} + \lambda_P\bm{F}) = \bm{0}$. Since $\bm{B}$ is invertible, the solution for this equation is $\bV^{(p)}(t) = -\lambda_P\bm{B}^{-1}\bm{F}$. The solution is $\bV(t) = \bV^{(h)}(t) + \bV^{(p)}(t)$. Applying the initial condition, $\bV(0) = \bm{C} -\lambda_P\bm{B}^{-1}\bm{F} = \bV^0$, we find $\bm{C} = \bV^0 + \lambda_P\bm{B}^{-1}\bm{F}$. Therefore, we have proved that the solution for the IVP problem in~(\ref{eq:odePD}) is indeed $\bV(t) = \displaystyle \mathrm{exp}(\frac{1}{\lambda_D + 1}\bm{B}t)(\bV^0 + \bm{B}^{-1}\bm{F}) - \lambda_P\bm{B}^{-1}\bm{F}.$\\ \\
In Section~\ref{subsec:PD-robust}, we show that $\bm{B}$ has only eigenvalues with negative real parts. As the result in Appendix~\ref{secapp:first-order-ode}, when $t \rightarrow 0$, the $\mathrm{exp}(\displaystyle\frac{1}{\lambda_D + 1}\bm{B}t) \rightarrow \bm{0}$ , leading to the vanishing of the $\bV^{(h)}(t)$. Hence the steady state solution for the ODE in~(\ref{eq:odePD}) becomes $ - \lambda_P\bm{B}^{-1}\bm{F}$. We have proved Lemma~\ref{lem:state-space-PD-sol}.
\subsection{Proof of Proposition~\ref{lem:lambda_PI}}
\label{secapp:lambda_PI}
Let
\begin{align}
    \bm{M} = \begin{bmatrix} \bm{0} & \bm{I} \\ \displaystyle -\frac{\lambda_I\bm{I}}{\lambda_D + 1} &\displaystyle \frac{\bm{K} - (\lambda_P + 1)\bm{I}}{\lambda_D + 1} \end{bmatrix}
\end{align}
For the solution of~(\ref{eq:ode-pi}) to be stable, the real part of eigenvalues of $\bm{M}$ must be all negative. Let $\bm{B} := \bm{K} - (\lambda_P + 1)\bm{I}$, for any eigenvalue $\gamma$ of $\bm{M}$
\begin{equation}
\label{eq:det-pi}
\begin{aligned}
\mathrm{det}( \bm{M} - \gamma\bm{I}) &= \mathrm{det}\Bigg( \begin{bmatrix} \bm{-\gamma\bm{I}} & \bm{I} \\ -\displaystyle \frac{\lambda_I}{\lambda_D + 1}\bm{I} & \displaystyle\frac{1}{\lambda_D + 1}(\bm{B} - \gamma\bm{I}) \end{bmatrix}\Bigg) \\
&= \mathrm{det}\Big(\frac{1}{\lambda_D + 1}(-\gamma\bm{B} + \gamma^2\bm{I} + \lambda_I \bm{I})\Big), &&\ (\text{since $\bm{B} - \gamma\bm{I}$ and $-\lambda_I\bm{I}$ are commute, see~\cite{0826aabe-2dc2-393a-bbf2-a98a36b6bed5}}) \\
&= 0\\
\end{aligned}
\end{equation}
Notice that $\gamma = 0$ is not a solution of~(\ref{eq:det-pi}). This fact is proved by contradiction. If $\gamma = 0$ is a solution, $\mathrm{det}(-\gamma\bm{B} + \gamma^2\bm{I} + \lambda_I\bm{I}) = \mathrm{det}(\lambda_I\bm{I}) = (\lambda_I)^N\mathrm{det}(\bm{I}) = (\lambda_I)^N > 0$ because $\lambda_I > 0$. This is contradict to~(\ref{eq:det-pi}). Since $\gamma \neq 0$, we can rewrite~(\ref{eq:det-pi}) as:
\begin{align}
    (-\frac{\gamma}{\lambda_D + 1})^N\mathrm{det}(\bm{B} - (\gamma + \frac{\lambda_I}{\gamma})\bm{I}) &=  0\\
    \iff \mathrm{det}(\bm{B} - (\gamma + \frac{\lambda_I}{\gamma})\bm{I}) &=  0.
\end{align}
Therefore, $\displaystyle \gamma + \frac{\lambda_I}{\gamma}$ are eigenvalues of $\bm{B}$. Given $\kappa_i$, for $i = 1, \dots, m; m \leq N$ are eigenvalues of $\bm{B}$. For each $i$, we find the solution of 
\begin{align}
\label{eq:quad-eigen}
    \gamma_i + \frac{\lambda_I}{\gamma_i} &= \kappa_i\\
    \iff \gamma_i^2 -\kappa \gamma_i + \lambda_I &= 0
\end{align}
Let $\gamma_{i, 1}, \gamma_{i, 1}$ are the solution of~(\ref{eq:quad-eigen}), and then 
\begin{equation}
\label{eq:condition-real}
    \begin{cases}
        \gamma_{i, 1} + \gamma_{i, 2} = \kappa_i\\
        \gamma_{i, 1}\gamma_{i, 2} = \lambda_I
    \end{cases}
    \iff \begin{cases}
        \mathrm{Re}(\gamma_{i, 1}) + \mathrm{Re}(\gamma_{i, 2}) = \mathrm{Re}(\kappa_i)\\
        \mathrm{Im}(\gamma_{i, 1}) + \mathrm{Im}(\gamma_{i, 2}) = \mathrm{Im}(\kappa_i)\\
        \mathrm{Re}(\gamma_{i, 1})\mathrm{Re}(\gamma_{i, 2}) - \mathrm{Im}(\gamma_{i, 1})\mathrm{Im}(\gamma_{i, 2}) = \lambda_I\\
        \mathrm{Re}(\gamma_{i, 1})\mathrm{Im}(\gamma_{i, 2}) + \mathrm{Im}(\gamma_{i, 1})\mathrm{Re}(\gamma_{i, 2}) = 0
    \end{cases}
\end{equation}
In Section~\ref{subsec:PD-robust}, we show that $\bm{B}$ has only eigenvalues with negative real parts. Hence, $\mathrm{Re}(\kappa_i) < 0$. 
Firstly, without any loss of generalization, suppose that $\mathrm{Re}(\gamma_{i, 1}) = 0$. This means
\begin{equation}
    \begin{cases}
        \mathrm{Re}(\gamma_{i, 2}) = \mathrm{Re}(\kappa_i) < 0\\
        - \mathrm{Im}(\gamma_{i, 1})\mathrm{Im}(\gamma_{i, 2}) = \lambda_I\\
        \mathrm{Im}(\gamma_{i, 1})\mathrm{Re}(\gamma_{i, 2}) = 0
    \end{cases}
    \Rightarrow \begin{cases}
        \mathrm{Im}(\gamma_{i, 1}) = 0\\
        - \mathrm{Im}(\gamma_{i, 1})\mathrm{Im}(\gamma_{i, 2}) = 0 \neq \lambda_I > 0
    \end{cases}
\end{equation}
which causes contradiction. Therefore, $\mathrm{Re}(\gamma_{i, 1}) \neq 0$. As the result, $\mathrm{Im}(\gamma_{i, 2}) =\displaystyle -\frac{\mathrm{Im}(\gamma_{i, 1})\mathrm{Re}(\gamma_{i, 2})}{\mathrm{Re}(\gamma_{i, 1})}$, substituting to~(\ref{eq:condition-real}), we obtain
\begin{align}
\label{eq:re1}
    \mathrm{Re}(\gamma_{i, 1})\mathrm{Re}(\gamma_{i, 2}) = \lambda_I - \mathrm{Im}(\gamma_{i, 1})^2\frac{\mathrm{Re}(\gamma_{i, 2})}{\mathrm{Re}(\gamma_{i, 1})}.
\end{align}
Suppose that $\mathrm{Re}(\gamma_{i, 1})\mathrm{Re}(\gamma_{i, 2}) < 0$, hence $\displaystyle \frac{\mathrm{Re}(\gamma_{i, 2})}{\mathrm{Re}(\gamma_{i, 1})} < 0$ leading to $\displaystyle - \mathrm{Im}(\gamma_{i, 1})^2\frac{\mathrm{Re}(\gamma_{i, 2})}{\mathrm{Re}(\gamma_{i, 1})} > 0$, (because $\mathrm{Im}(\gamma_{i, 1})^2 > 0$). Therefore the RHS of~(\ref{eq:re1}) is greater than 0 (since $\lambda_I$ also greater than $0$), which contradicts our assumption that $\mathrm{Re}(\gamma_{i, 1})\mathrm{Re}(\gamma_{i, 2}) < 0$. As a consequence, we obattain the following result:
\begin{equation}
    \begin{cases}
        \mathrm{Re}(\gamma_{i, 1}) + \mathrm{Re}(\gamma_{i, 2}) = \mathrm{Re}(\kappa_i) < 0\\
        \mathrm{Re}(\gamma_{i, 1})\mathrm{Re}(\gamma_{i, 2}) > 0
    \end{cases}
    \iff \begin{cases}
        \mathrm{Re}(\gamma_{i, 1}) < 0\\
        \mathrm{Re}(\gamma_{i, 2}) < 0,
    \end{cases}
\end{equation}
for $i = 1, \dots, m$. Therefore, all eigenvalues of $\bm{M}$ as negative real parts. Combined with result in Appendix~\ref{secapp:first-order-ode}, we have the system described by~(\ref{eq:ode-pi}) has stable solution when $t \rightarrow 0$, for all $\lambda_P, \lambda_I, \lambda_D > 0$. This concludes our proof.
\subsection{The Fretchet derivation of the derivative of $J$ w.r.t $v_j$.}
The partial derivative ${\partial J}/{\partial v_j}$, $j = 1, 2, \dots, D$, is defined through its dot product with an arbitrary function $h_j \in L^{2}(\Omega \times [0, \infty))$ as follows

\begin{align}  
    \frac{\partial J}{\partial v_j}\cdot h_j(x, t) &= \frac{d}{d\tau}J(v_j + \tau h_j)\bigl|_{\tau=0} \nonumber \\ 
&= \frac{1}{2} \left(\frac{d}{d\tau}  \int_{\Omega \times \Omega}(v_j(x) - v_j(y) + \tau h_j(x) - \tau h_j(y))^{2} k(x, y)dxdy\right)\biggl|_{\tau=0} \nonumber \\
&= \left(\int_{\Omega \times \Omega}(v_j(x, t) - v_j(y) + \tau h_j(x) - \tau h_j(y, t))(h_j(x) -h_j(y))k(x, y)dxdy\right)\biggl|_{\tau=0} \nonumber \\
&= \int_{\Omega \times \Omega}(v_j(x) - v_j(y))(h_j(x) -h_j(y))k(x, y)dxdy \nonumber \\
&= \int_{\Omega \times \Omega}(v_j(x) - v_j(y))h_j(x)k(x, y)dxdy - \int_{\Omega \times \Omega}(v_j(x) - v_j(y))h_j(y)k(x, y)dxdy \nonumber
\end{align}
Applying a change of variables $(x, y) \rightarrow (y, x)$ to the second term of the above integral, we have
\begin{align}  
\frac{\partial J}{\partial v_j} \cdot h_j(x) &= \int_{\Omega \times \Omega}(v_j(x) - v_j(y))h_j(x)k(x, y)dxdy - \int_{\Omega \times \Omega}(v_j(y) - v_j(x))h_j(x, t)k(y, x)dxdy \nonumber \\
&= \int_{\Omega \times \Omega}(v_j(x) - v_j(y)(k(x, y) + k(y, x))dyh_j(x)dx \nonumber
\end{align}
Thus, the Frechet derivative of J with respect to $v_j$ is given by
\begin{align}  
\nonumber
\frac{\partial J}{\partial v_j} 
&= \int_{\Omega}(v_j(x) - v_j(y)(k(x, y) + k(y, x))dy. 
\end{align}
\subsection{The derivation of the gradient flow of $
 E(v, f)$}
\label{secapp:derE}
 Taking the gradient of $E(\bv, \bff)$ with respect to $\bv$, we obtain
\label{secapp:djv}
\begin{align}
\label{eq:devEu}
    \nabla_{\bm{v}}E = \nabla_{\bm{v}}J + 
    \left[\frac{\partial G}{\partial u_1}, \frac{\partial G}{\partial u_2}, \dots, \frac{\partial G}{\partial u_{D}} \right]^T.
\end{align}
The partial derivative ${\partial G}/{\partial v_j}$, $j = 1, 2, \dots, D$, is defined through its dot product with an arbitrary function $h_j \in L^{2}(\Omega)$ as follows
\begin{align}
\frac{\partial G}{\partial v_j}\cdot h_j(x) &= \frac{d}{d\tau}G(v_j + \tau h_j)\bigl|_{\tau=0} \nonumber\\
&= \frac{\lambda}{2} \left(\frac{d}{d\tau} \int_{\Omega} (v_j(x) - f_j(x) + \tau h_j(x))^2 dx \right)\biggl|_{\tau=0}  \nonumber\\
&= \lambda \int_{\Omega} (v_j(x) - f_j(x))h_j(x)  dx  \nonumber.
\end{align}
Thus, the Frechet derivative of F with respect to $v_j$ is given by
\begin{equation}
\label{eq:partial-devFu}
\frac{\partial G}{\partial v_j} = \lambda (v_j(x) - f_j(x))
\end{equation}
Substituting the formula for ${\partial G}/{\partial v_j}$ in~(\ref{eq:partial-devFu}) into~(\ref{eq:devEu}) for $\nabla_{\bv}E(\bv, \bff)$, we obtain the following gradient flow 
\begin{equation}
    \label{eq:gradient-descent-E}
    \frac{d\bm{v}(x,t)}{dt} = -\nabla_{\bm{v}}E(\bv, \bff) = -\nabla_{\bm{v}}J(\bv)(x) + \lambda \bigl(\bff(x) - \bv(x)\bigl).
\end{equation}
\\
This concludes the derivation.
\subsection{The derivation of~(\ref{eq:dis_breg})}
\label{secapp:grad-flow-breg}
Denote $H(\bv, \bff) :=\displaystyle \frac{\lambda}{2} \int_{\Omega} \|\bv(x) - \bff(x) - \bm{e}^{\ell}(x)\|_2^2 dx$. 
 Taking the gradient of $J(\bv) + H(\bv, \bff)$ with respect to $\bv$, we obtain
\begin{align}
\label{eq:devJHu}
    \nabla_{\bm{v}}E = \nabla_{\bm{v}}J + 
    \left[\frac{\partial H}{\partial v_1}, \frac{\partial H}{\partial v_2}, \dots, \frac{\partial H}{\partial v_{D}} \right]^T.
\end{align}
The partial derivative ${\partial H}/{\partial v_j}$, $j = 1, 2, \dots, D$, is defined through its dot product with an arbitrary function $h_j \in L^{2}(\Omega)$ as follows
\begin{align}
\frac{\partial H}{\partial v_j}\cdot h_j(x) &= \frac{d}{d\tau}H(v_j + \tau h_j)\bigl|_{\tau=0} \nonumber\\
&= \frac{\lambda}{2} \left(\frac{d}{d\tau} \int_{\Omega} (v_j(x) - f_j(x) - e^{\ell}_j(x) + \tau h_j(x))^2 dx \right)\biggl|_{\tau=0}  \nonumber\\
&= \lambda \int_{\Omega} (v_j(x) - f_j(x) -e^{\ell}_j)h_j(x)  dx  \nonumber.
\end{align}
Thus, the Frechet derivative of F with respect to $v_j$ is given by
\begin{equation}
\label{eq:partial-devHu}
\frac{\partial H}{\partial v_j} = \lambda (v_j(x) - f_j(x) - e^{\ell}_j)
\end{equation}
Substituting the formula for ${\partial H}/{\partial v_j}$ in~(\ref{eq:partial-devHu}) into~(\ref{eq:devJHu}) for $\nabla_{\bv}E(\bv, \bff)$, we obtain the following gradient flow 
at iteration $\ell + 1$ 
\begin{equation}
\begin{aligned}
   \label{eq:grad-flow-breg}
        \displaystyle\frac{d\bv(x, t)}{dt} &= \int_{\Omega} \bigl(\bm{v}(y,t) - \bm{v}(x,t) \bigl)\bigl(k(x, y) + k(y,x)\bigl)dy \\
        &+ \lambda \bigl(\bff(x) - \bv(x, t) + \bm{e}^{\ell}(x)\bigl).
\end{aligned}
\end{equation}
Applying Euler method to discretize~(\ref{eq:grad-flow-breg}) with $\Delta t = 1$ and $\bv(x, 0) = \bv^{\ell}(x)$, we approximate the $\bv^{\ell + 1}$ with one-step gradient descent:
\begin{equation}
\begin{aligned}
\nonumber
   \bv^{\ell + 1}(x) 
   &= \int_{\Omega} \bigl(\bm{v}^{\ell}(y) - \bm{v}^{\ell}(x) \bigl)\bigl(k(x, y) + k(y,x)\bigl)dy \\
   & + \bv^{\ell}(x) + \lambda \bm{e}^{\ell}(x) + \lambda\bm{e}^{\ell}_{\mathrm{a}}(x).
\end{aligned}
\end{equation}
This concludes the derivation.
\section{Additional Experiment results}
\subsection{PID DeiT and softmax DeiT under escalating perturbation attacks.}
\begin{figure}[!t]
\centering
\includegraphics[width=0.9\textwidth]{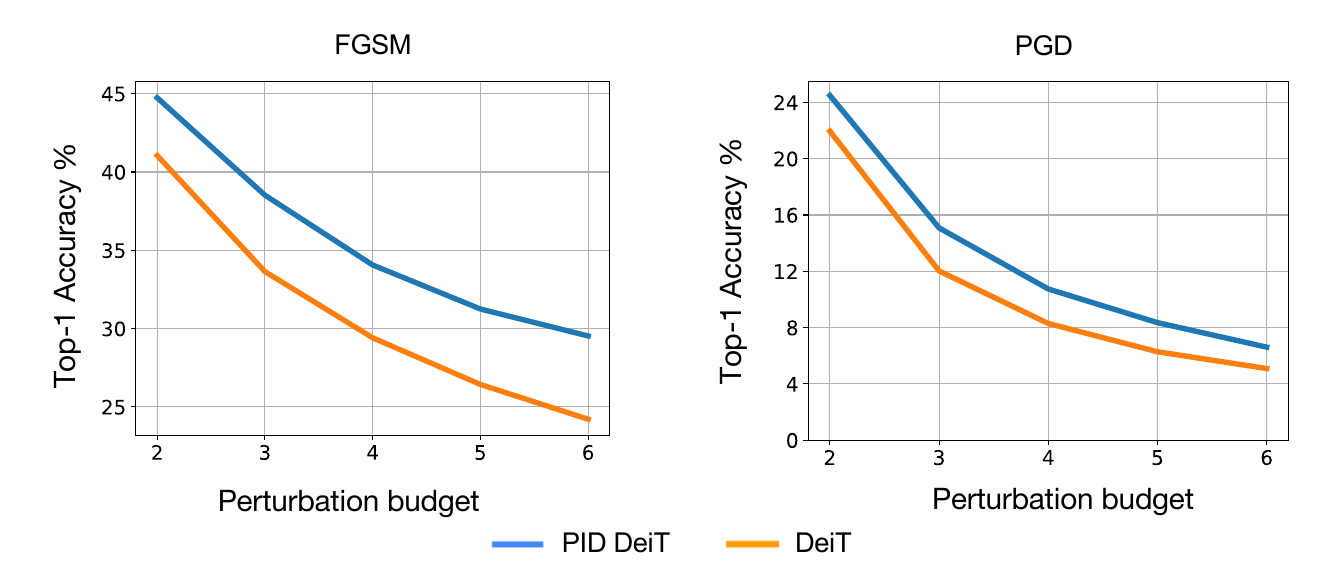}
\caption{\small The top-1 classification accuracy curves on ImageNet against FGSM and PGD attack methods, plotted against perturbation budgets (scaled by 255).}
\label{fig:pid-deit-more-attacks} 
\end{figure}

We evaluate PID DeiT and softmax DeiT under FGSM and PGD attack methods with increasing perturbation budgets (see Fig.~\ref{fig:pid-deit-more-attacks}) (scaled by 255). The proposed PID DeiT exhibits stronger defense in both attack methods and various perturbation budgets.
\label{secapp:moreattacks}


\end{document}